\documentclass[10pt,conference,compsoc]{IEEEtran}
\newcommand{\etal}{{\emph{et~al.}}~}
\newcommand{\eg}{{\emph{e.g.}}~}

\usepackage{graphicx}
\usepackage{amsmath,amssymb}

\ifCLASSOPTIONcompsoc
  \usepackage[nocompress]{cite}
\else
  \usepackage{cite}
\fi
\ifCLASSINFOpdf
\else
\fi
\begin{document}
%
% paper title
% Titles are generally capitalized except for words such as a, an, and, as,
% at, but, by, for, in, nor, of, on, or, the, to and up, which are usually
% not capitalized unless they are the first or last word of the title.
% Linebreaks \\ can be used within to get better formatting as desired.
% Do not put math or special symbols in the title.
\title{Skeleton-Based Action Recognition Using \\ Spatio-Temporal LSTM Network with Trust Gates}
%
%
% author names and IEEE memberships
% note positions of commas and nonbreaking spaces ( ~ ) LaTeX will not break
% a structure at a ~ so this keeps an author's name from being broken across
% two lines.
% use \thanks{} to gain access to the first footnote area
% a separate \thanks must be used for each paragraph as LaTeX2e's \thanks
% was not built to handle multiple paragraphs
%
%
%\IEEEcompsocitemizethanks is a special \thanks that produces the bulleted
% lists the Computer Society journals use for "first footnote" author
% affiliations. Use \IEEEcompsocthanksitem which works much like \item
% for each affiliation group. When not in compsoc mode,
% \IEEEcompsocitemizethanks becomes like \thanks and
% \IEEEcompsocthanksitem becomes a line break with idention. This
% facilitates dual compilation, although admittedly the differences in the
% desired content of \author between the different types of papers makes a
% one-size-fits-all approach a daunting prospect. For instance, compsoc
% journal papers have the author affiliations above the "Manuscript
% received ..."  text while in non-compsoc journals this is reversed. Sigh.

\author{Jun~Liu, Amir~Shahroudy, Dong~Xu, Alex~C.~Kot, and~Gang~Wang
% <-this % stops a space
\IEEEcompsocitemizethanks{
\IEEEcompsocthanksitem J. Liu, A. Shahroudy, and A. C. Kot
are with School of Electrical and Electronic Engineering, Nanyang Technological University, Singapore.\protect\\
E-mail: \{jliu029, amir3, eackot\}@ntu.edu.sg.
\IEEEcompsocthanksitem G. Wang is with Alibaba Group, Hangzhou, 310052, China.\protect\\
E-mail: wanggang@ntu.edu.sg.
\IEEEcompsocthanksitem D. Xu is with School of Electrical and Information Engineering,
University of Sydney, Sydney, NSW 2006, Australia.\protect\\
E-mail: dong.xu@sydney.edu.au.}% <-this % stops an unwanted space
\thanks{Manuscript received August, 2016; revised June, 2017.}}

\IEEEtitleabstractindextext{%
\begin{abstract}
Skeleton-based human action recognition has attracted a lot of research attention during the past few years.
Recent works attempted to utilize recurrent neural networks to model the temporal dependencies between the 3D positional configurations of human body joints for better analysis of human activities in the skeletal data.
The proposed work extends this idea to spatial domain as well as temporal domain to better analyze the hidden sources of action-related information within the human skeleton sequences in both of these domains simultaneously.
Based on the pictorial structure of Kinect's skeletal data, an effective tree-structure based traversal framework is also proposed.
In order to deal with the noise in the skeletal data, a new gating mechanism within LSTM module is introduced,
with which the network can learn the reliability of the sequential data
and accordingly adjust the effect of the input data on the updating procedure of the long-term context representation stored in the unit's memory cell.
Moreover, we introduce a novel multi-modal feature fusion strategy within the LSTM unit in this paper.
The comprehensive experimental results on seven challenging benchmark datasets for human action recognition demonstrate the effectiveness of the proposed method.
\end{abstract}

% Note that keywords are not normally used for peerreview papers.
}

% make the title area
\maketitle

% To allow for easy dual compilation without having to reenter the
% abstract/keywords data, the \IEEEtitleabstractindextext text will
% not be used in maketitle, but will appear (i.e., to be "transported")
% here as \IEEEdisplaynontitleabstractindextext when the compsoc
% or transmag modes are not selected <OR> if conference mode is selected
% - because all conference papers position the abstract like regular
% papers do.
\IEEEdisplaynontitleabstractindextext
% \IEEEdisplaynontitleabstractindextext has no effect when using
% compsoc or transmag under a non-conference mode.

% For peer review papers, you can put extra information on the cover
% page as needed:
% \ifCLASSOPTIONpeerreview
% \begin{center} \bfseries EDICS Category: 3-BBND \end{center}
% \fi
%
% For peerreview papers, this IEEEtran command inserts a page break and
% creates the second title. It will be ignored for other modes.
\IEEEpeerreviewmaketitle

\vspace{1.5cm}

\IEEEraisesectionheading{
\section{Introduction}\label{sec:intro}}
% Computer Society journal (but not conference!) papers do something unusual
% with the very first section heading (almost always called "Introduction").
% They place it ABOVE the main text! IEEEtran.cls does not automatically do
% this for you, but you can achieve this effect with the provided
% \IEEEraisesectionheading{} command. Note the need to keep any \label that
% is to refer to the section immediately after \section in the above as
% \IEEEraisesectionheading puts \section within a raised box.

% The very first letter is a 2 line initial drop letter followed
% by the rest of the first word in caps (small caps for compsoc).
%
% form to use if the first word consists of a single letter:
% \IEEEPARstart{A}{demo} file is ....
%
% form to use if you need the single drop letter followed by
% normal text (unknown if ever used by the IEEE):
% \IEEEPARstart{A}{}demo file is ....
%
% Some journals put the first two words in caps:
% \IEEEPARstart{T}{his demo} file is ....
%
% Here we have the typical use of a "T" for an initial drop letter
% and "HIS" in caps to complete the first word.
\IEEEPARstart{H}{uman} action recognition is a fast developing research area due to its wide applications
in intelligent surveillance, human-computer interaction, robotics, and so on.
In recent years, human activity analysis based on human skeletal data has attracted a lot of attention,
%Having a reliable set of 3D positions of important body joints, we can apply different analysis approaches to classify the activities done by human subjects.
and various methods for feature extraction and classifier learning have been developed for skeleton-based action recognition \cite{zhu2016handcrafted,presti20163d,han2016review}.
A hidden Markov model (HMM) is utilized by Xia \etal \cite{HOJ3D} to model the temporal dynamics over a histogram-based representation of joint positions for action recognition.
The static postures and dynamics of the motion patterns are represented via eigenjoints by Yang and Tian \cite{eigenjointsJournal}.
A Naive-Bayes-Nearest-Neighbor classifier learning approach is also used by \cite{eigenjointsJournal}.
Vemulapalli \etal \cite{vemulapalli2014liegroup} represent the skeleton configurations and action patterns as points and curves in a Lie group,
and then a SVM classifier is adopted to classify the actions.
Evangelidis \etal \cite{skeletalQuads} propose to learn a GMM over the Fisher kernel representation of the skeletal quads feature.
An angular body configuration representation over the tree-structured set of joints is proposed in \cite{hog2-ohnbar}. %which calculates the similarity of these features over temporal dimension to build the global representation of the action samples and fed them to SVM for final classification.
A skeleton-based dictionary learning method using geometry constraint and group sparsity is also introduced in \cite{Luo_2013_ICCV}.

Recently, recurrent neural networks (RNNs) which can handle the sequential data with variable lengths \cite{graves2013speechICASSP,sutskever2014sequence},
have shown their strength in language modeling \cite{mikolov2011extensions,sundermeyer2012lstm,mesnil2013investigation},
image captioning \cite{vinyals2015show,xu2015show},
video analysis \cite{srivastava2015unsupervised,Singh_2016_CVPR,Jain_2016_CVPR,Alahi_2016_CVPR,Deng_2016_CVPR,Ibrahim_2016_CVPR,Ma_2016_CVPR,Ni_2016_CVPR,li2016online},
%human re-identification \cite{varior2016siamese,varior2016learning},
and RGB-based activity recognition \cite{yue2015beyond,donahue2015long,li2016action,wu2015ACMMM}.
Applications of these networks have also shown promising achievements in skeleton-based action recognition \cite{du2015hierarchical,veeriah2015differential,nturgbd}.

In the current skeleton-based action recognition literature, RNNs are mainly used to model the long-term context information across the temporal dimension by representing motion-based dynamics.
However, there is often strong dependency relations among the skeletal joints in spatial domain also,
and the spatial dependency structure is usually discriminative for action classification.

%Current RNN based 3D action recognition methods utilized the recurrence learning on the temporal domain in order to discover the dynamics of body poses in time.
%However, there is another essential source of dependency among the 3D skeleton data, which relates the locations of individual joints of the body together.
%3D pose of the skeleton on each video frame can also be considered as a sequence of the input data and since they are highly dependent and overall are informative about the class of the underlying action, it is intuitive to analyse them in a better way.
%The downside of most of the current methods is to ignore these dependencies and concatenate the input data from all the joints together to form the input of the recurrent network at each frame.

To model the dynamics and dependency relations in both temporal and spatial domains,
we propose a spatio-temporal long short-term memory (ST-LSTM) network in this paper.
%Different strategies are studied in order to find the best and most representative design of the sequential information in spatial domain.
In our ST-LSTM network,
each joint can receive context information from its stored data from previous frames and also from the neighboring joints at the same time frame to represent its incoming spatio-temporal context.
Feeding a simple chain of joints to a sequence learner limits the performance of the network,
as the human skeletal joints are not semantically arranged as a chain.
Instead, the adjacency configuration of the joints in the skeletal data can be better represented by a tree structure.
Consequently, we propose a traversal procedure by following the tree structure of the skeleton
to exploit the kinematic relationship among the body joints for better modeling spatial dependencies.

Since the 3D positions of skeletal joints provided by depth sensors are not always very accurate,
we further introduce a new gating framework, so called ``trust gate'',
for our ST-LSTM network to analyze the reliability of the input data at each spatio-temporal step.
The proposed trust gate gives better insight to the ST-LSTM network about
when and how to update, forget, or remember the internal memory content as the representation of the long-term context information.

In addition, we introduce a feature fusion method within the ST-LSTM unit to better exploit the multi-modal features extracted for each joint.

We summarize the main contributions of this paper as follows.
%\begin{enumerate}
(1) A novel spatio-temporal LSTM (ST-LSTM) network for skeleton-based action recognition is designed.
(2) A tree traversal technique is proposed to feed the structured human skeletal data into a sequential LSTM network.
(3) The functionality of the ST-LSTM framework is further extended by adding the proposed ``trust gate''.
(4) A multi-modal feature fusion strategy within the ST-LSTM unit is introduced.
(5) The proposed method achieves state-of-the-art performance on seven benchmark datasets.
%\end{enumerate}

The remainder of this paper is organized as follows.
In section \ref{sec:relatedwork}, we introduce the related works on skeleton-based action recognition, which used recurrent neural networks to model the temporal dynamics.
In section \ref{sec:approach}, we introduce our end-to-end trainable spatio-temporal recurrent neural network for action recognition.
The experiments are presented in section \ref{sec:exp}.
Finally, the paper is concluded in section \ref{sec:conclusion}.

\section{Related Work}
\label{sec:relatedwork}

Skeleton-based action recognition has been explored in different aspects during recent years \cite{7284883,actionletPAMI,MMMP_PAMI,MMTW,Vemulapalli_2016_CVPR,rahmani2014real,shahroudy2014multi,rahmani2015learning,lillo2014discriminative,
%hu2016ECCV,
%shahroudy2016deep,
%du2016representation,
%rahmani20163d,
jhuang2013towards,
chen_2016_icassp,liu2016IVC,cai2016TMM,al2016PRL,Tao_2015_ICCV_Workshops
}.
In this section, we limit our review  to more recent approaches which use RNNs or LSTMs for human activity analysis.% which are more related to the proposed framework.

Du \etal \cite{du2015hierarchical} proposed a Hierarchical RNN network by utilizing multiple bidirectional RNNs in a novel hierarchical fashion.
The human skeletal structure was divided to five major joint groups.
Then each group was fed into the corresponding bidirectional RNN.
The outputs of the RNNs were concatenated to represent the upper body and lower body,
then each was further fed into another set of RNNs.
By concatenating the outputs of two RNNs, the global body representation was obtained, which was fed to the next RNN layer.
Finally, a softmax classifier was used in \cite{du2015hierarchical} to perform action classification.

Veeriah \etal \cite{veeriah2015differential} proposed to add a new gating mechanism for LSTM to model the derivatives of the memory states and explore the salient action patterns.
In this method, all of the input features were concatenated at each frame and were fed to the differential LSTM at each step.

Zhu \etal \cite{zhu2016co} introduced a regularization term to the objective function of the LSTM
network to push the entire framework towards learning co-occurrence relations among the joints for action recognition.
An internal dropout \cite{dropout} technique within the LSTM unit was also introduced in \cite{zhu2016co}.

Shahroudy \etal \cite{nturgbd} proposed to split the LSTM's memory cell to sub-cells to push the network towards learning the context representations for each body part separately.
The output of the network was learned by concatenating the multiple memory sub-cells.

Harvey and Pal \cite{harvey2015semi} adopted an encoder-decoder recurrent network to reconstruct the skeleton sequence and perform action classification at the same time.
Their model showed promising results on motion capture sequences.

Mahasseni and Todorovic \cite{mahasseni2016regularizing} proposed to use LSTM to encode a skeleton sequence as a feature vector.
At each step, the input of the LSTM consists of the concatenation of the skeletal joints' 3D locations in a frame.
They further constructed a feature manifold by using a set of encoded feature vectors.
Finally, the manifold was used to assist and regularize the supervised learning of another LSTM for RGB video based action recognition.

Different from the aforementioned works,
our proposed method does not simply concatenate the joint-based input features to build the body-level feature representation.
Instead, the dependencies between the skeletal joints are explicitly modeled by applying recurrent analysis over temporal and spatial dimensions concurrently.
Furthermore, a novel trust gate is introduced to make our ST-LSTM network more reliable against the noisy input data.

%A preliminary version of this work has been presented in \cite{liu2016spatio}, in which we mainly validated the effectiveness of our model on four benchmark datasets.
This paper is an extension of our preliminary conference version \cite{liu2016spatio}.
In \cite{liu2016spatio}, we validated the effectiveness of our model on four benchmark datasets.
%In this paper, we further extend our model to handle the action recognition problem in the more challenging RGB videos, in which the 2D skeletal joints are extracted by using a pose estimation algorithm.
%to handle the action recognition problem in the more challenging RGB videos, in which the 2D skeletal joints are extracted by using a pose estimation algorithm.
In this paper, we extensively evaluate our model on seven challenging datasets.
Besides, we further propose an effective feature fusion strategy inside the ST-LSTM unit.
In order to improve the learning ability of our ST-LSTM network, a last-to-first link scheme is also introduced.
In addition, we provide more empirical analysis of the proposed framework.

\section{Spatio-Temporal Recurrent Networks}
\label{sec:approach}

%Talk about the problem and notations... How is the structure of the data. Where it comes from. What we want to do to it?!?! Introduce the baseline LSTM on skeleton framework for action recognition.

%Human actions can be represented by the motion of body parts over time.
In a generic skeleton-based action recognition problem, the input observations are limited to the 3D locations of the major body joints at each frame.
Recurrent neural networks have been successfully applied to %%RGB-based human activity analysis \cite{srivastava2015icml,donahue2015long,veeriah2015differential} and 3D action recognition \cite{du2015hierarchical,zhu2016co,nturgbd}.
this problem recently \cite{du2015hierarchical,zhu2016co,nturgbd}.
LSTM networks \cite{lstm} are among the most successful extensions of recurrent neural networks.
A gating mechanism controlling the contents of an internal memory cell is adopted by the LSTM model
to learn a better and more complex representation of long-term dependencies in the input sequential data.
Consequently, LSTM networks are very suitable for feature learning over time series data (such as human skeletal sequences over time).
%The core idea behind the model is a built-in memory cell that stores information over time to explore long range dynamics, with nonlinear gate units controlling the information flow into and out of the cell \cite{wu2015modeling}.

We will briefly review the original LSTM model in this section,
and then introduce our ST-LSTM network and the tree-structure based traversal approach.
We will also introduce a new gating mechanism for ST-LSTM to handle the noisy measurements in the input data for better action recognition.
Finally, an internal feature fusion strategy for ST-LSTM will be proposed.

\subsection{Temporal Modeling with LSTM}
\label{sec:approach:lstm}

In the standard LSTM model, each recurrent unit contains an input gate $i_t$, a forget gate $f_t$, an output gate $o_t$, and an internal memory cell state $c_t$, together with a hidden state $h_t$.
The input gate $i_{t}$ controls the contributions of the newly arrived input data at time step $t$ for updating the memory cell,
while the forget gate $f_{t}$ determines how much the contents of the previous state $(c_{t-1})$ contribute to deriving the current state $(c_{t})$. %no need to cite here \cite{veeriah2015differential}.
The output gate $o_{t}$ learns how the output of the LSTM unit at current time step should be derived from the current state of the internal memory cell.
%Readers are referred to \cite{graves2012supervised} for more details about the mechanism of LSTM.
These gates and states can be obtained as follows:

\begin{eqnarray}
\left(
   \begin{array}{ccc}
    i_{t} \\
    f_{t} \\
    o_{t} \\
    u_{t} \\
   \end{array}
\right)
&=&
\left(
   \begin{array}{ccc}
    \sigma \\
    \sigma \\
    \sigma \\
    \tanh \\
   \end{array}
\right)
\left(
   M
   \left(
       \begin{array}{ccc}
        x_{t} \\
        h_{t-1} \\
       \end{array}
   \right)
\right)\\
c_{t} &=& i_{t} \odot u_{t}  + f_{t} \odot  c_{t-1}
\label{eq:ct}\\
h_{t} &=& o_{t}  \odot \tanh( c_{t})
\label{eq:ht}
\end{eqnarray}
where $x_t$ is the input at time step $t$, $u_t$ is the modulated input, $\odot$ denotes the element-wise product,
%$\sigma$ represents the sigmoid activation function.
%$M: \Re^A \to \Re^B$ is an affine transformation consisting of model parameters with $A = D + d$ and $B = 4d$, where $D$ is the dimensionality of $x_t$ and $d$ is the dimensionality of all of $i_t$, $f_t$, $c_t$, $o_t$, $h_t$, and $u_t$.
and $M: \mathbb{R}^{D+d} \to \mathbb{R}^{4d}$ is an affine transformation.
$d$ is the size of the internal memory cell, and $D$ is the dimension of $x_t$.
%Sigmoid $\sigma(\cdot)$ is the activation function with the range $[0, 1]$, and the range of $\tanh (\cdot)$ is [-1,1]. Therefore, the entries of the gating vectors $i_t$, $f_t$, and $o_t$ are all in $[0, 1]$, and $0$ means the gate is closed while $1$ represents gate is completely open.
%% Above commented line is too much of details, not needed ;)

\subsection{Spatio-Temporal LSTM}
\label{sec:approach:stlstm}
%What is missing in the temporal LSTM?
%
%Talk about our ST-LSTM is our solution to this problem.
%
%Give the specific and technical details; equations; figures; ...
%Here we have to clearly mention the time segments technique following NTU-dataset paper! :)

%Spatial-temporal discriminative information is very important to action recognition. LSTM has very good modeling ability of the temporal structure, thus it is powerful for action recognition over the frames. However, the spatial structure of the human pose in each frame is also necessary for accurate action recognition. Du et al. \cite{du2015hierarchical} proposed to use RNN to model the temporal structure in the video sequence. To employee the spatial structure, they proposed to decompose the human skeleton into five parts (two arms, two legs, and one trunk). Using hierarchical recurrent neural network to combine the five parts based on the physical configuration of the body, their model can be considered extracting spatial and temporal features of the videos. In this section, we will introduce a spatial-temporal LSTM (ST-LSTM) structure to take advantage of both spatial and temporal discriminative information for action recognition in skeleton sequences.

\begin{figure}
	\begin{minipage}[b]{1.0\linewidth}
		\centering
		\centerline{\includegraphics[scale=.338]{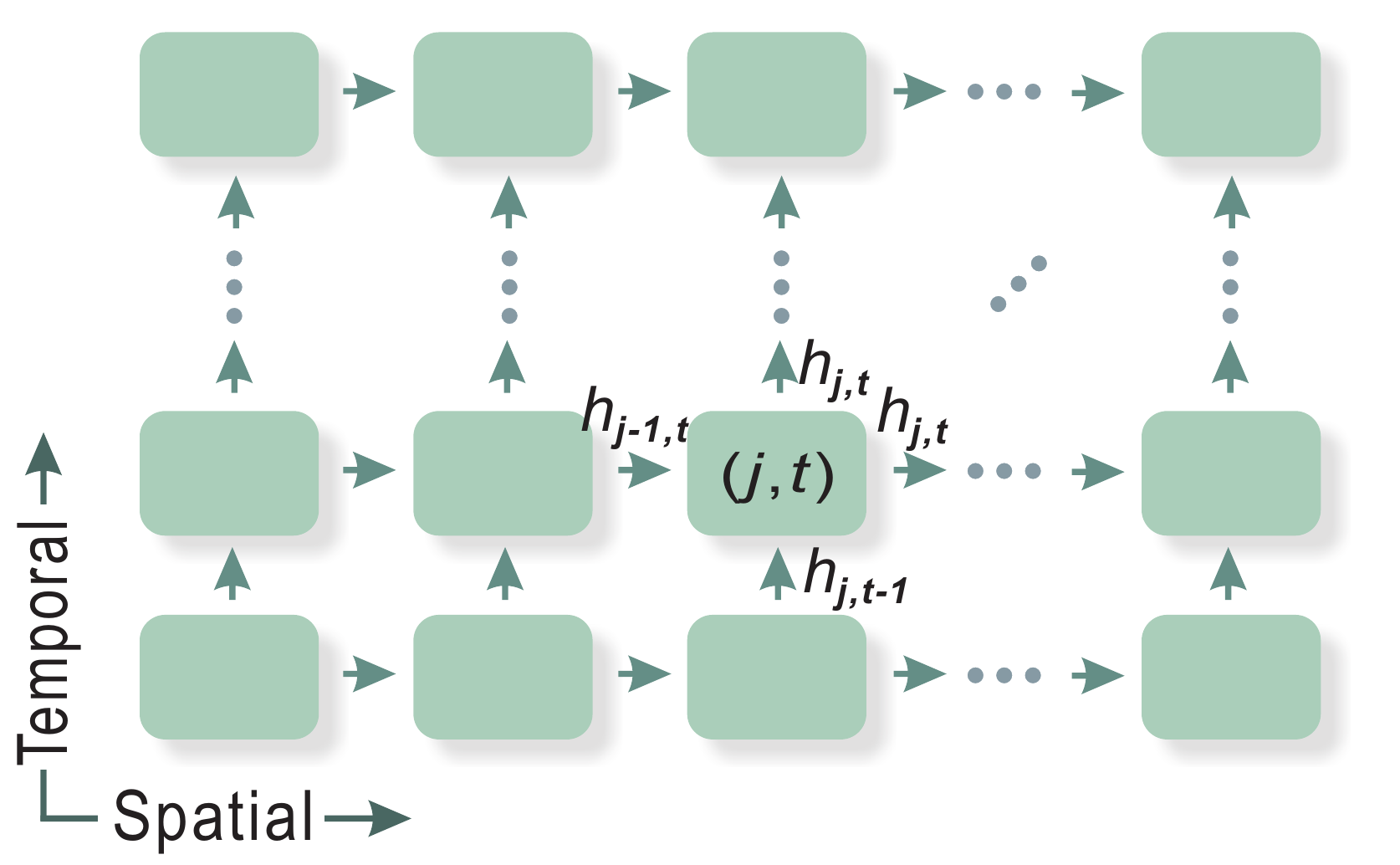}}
	\end{minipage}
	%\begin{minipage}[b]{0.5\linewidth}
	%  \centering
	%  \centerline{\includegraphics[scale=.3]{StackedSTLSTM.pdf}}
	%\end{minipage}
	\caption{
		Illustration of the spatio-temporal LSTM network.
        In temporal dimension, the corresponding body joints are fed over the frames.
		In spatial dimension, the skeletal joints in each frame are fed as a sequence. % the joints are fed to the network one by one by following the head - trunk - left arm - right arm - left leg - right leg order (refer to \figurename{ \ref{fig:tree16joints}}(a), the joint chain order is 3-2-1-10-4-5-6-7-8-9-11-12-13-14-15-16).
		Each unit receives the hidden representation of the previous joints and the same joint from previous frames.}
	%%ST-LSTM. In the spatial dimension, the joints are fed to the network one by one by following the head - trunk - left arm - right arm - left leg - right leg order (corresponding to \figurename{ \ref{fig:tree16joints}(a)}, the joint chain order is 3-2-1-10-4-5-6-7-8-9-11-12-13-14-15-16).}
	\label{fig:STLSTM}
\end{figure}

RNNs have already shown their strengths in modeling the complex dynamics of human activities as time series data,
and achieved promising performance in skeleton-based human action recognition \cite{du2015hierarchical,zhu2016co,veeriah2015differential,nturgbd}.
%, which proves the strengths of RNNs in modeling the complex dynamics of the human actions in temporal space.
%However, in the existing RNN-based methods for 3D action analysis, the focus were more on the temporal dynamics of the body poses, in which the RNNs or LSTMs are applied.
%Since the 3D body pose representation at each frame is also a sequence of joint locations in three dimensional space, it is reasonable to adopt a recurrent framework in the spatial level also, instead of feeding all the information of each frame to the temporal RNN.
%Based on this intuition we propose a spatio-temporal LSTM (ST-LSTM) structure, in which a new direction of information flow is added to the LSTM chain.
%In the spatial direction, we have a sequence of body joints and on the temporal dimension we have the unwrapped sequence of LSTM units over time.
%Each LSTM unit receives the hidden representation of the previous joints in the sequence and its own hidden representation in previous frames (\figurename{ \ref{fig:STLSTM}}).
%To analyse the input data in both spatial and temporal directions, and inspired from the tree-structured LSTM \cite{tai2015improved}, we extended the LSTM mechanism to be able to handle the long-term behavior of the input representation of the human actions.
In the existing literature, RNNs are mainly utilized in temporal domain to discover the discriminative dynamics and motion patterns for action recognition.
However, there is also discriminative spatial information encoded in the joints' locations and posture configurations at each video frame,
and the sequential nature of the body joints makes it possible to apply RNN-based modeling to spatial domain as well.

Different from the existing methods which concatenate the joints' information as the entire body's representation,
we extend the recurrent analysis to spatial domain by discovering the spatial dependency patterns among different body joints. % and to analyse these static posture patterns.
%, \emph{i.e.} the effect of other joints on the location of each individual joint. %due to the strong dependencies between body parts in each human body pose configuration.
%The spatial configuration of joints is highly discriminative for action recognition.
%Current methods concatenated the input features of the joints together, hence did not model the above-mentioned spatial dependency explicitly.
We propose a spatio-temporal LSTM (ST-LSTM) network to simultaneously model the temporal dependencies among different frames and also the spatial dependencies of different joints at the same frame.
Each ST-LSTM unit, which corresponds to one of the body joints,
receives the hidden representation of its own joint from the previous time step
and also the hidden representation of its previous joint at the current frame.
A schema of this model is illustrated in \figurename{ \ref{fig:STLSTM}}.

In this section, we assume the joints are arranged in a simple chain sequence, and the order is depicted in \figurename{ \ref{fig:tree16joints}(a)}.
In section \ref{sec:approach:skeltree}, we will introduce a more advanced traversal scheme to take advantage of the adjacency structure among the skeletal joints.

We use $j$ and $t$ to respectively denote the indices of joints and frames,
where $j \in \{1,...,J\}$ and $t \in \{1,...,T\}$.
Each ST-LSTM unit is fed with the input ($x_{j, t}$, the information of the corresponding joint at current time step),
the hidden representation of the previous joint at current time step $(h_{j-1,t})$,
and the hidden representation of the same joint at the previous time step $(h_{j,t-1})$.

\begin{figure}
	%\begin{minipage}[b]{0.5\linewidth}
	% \centering
	\centerline{\includegraphics[scale=0.479]{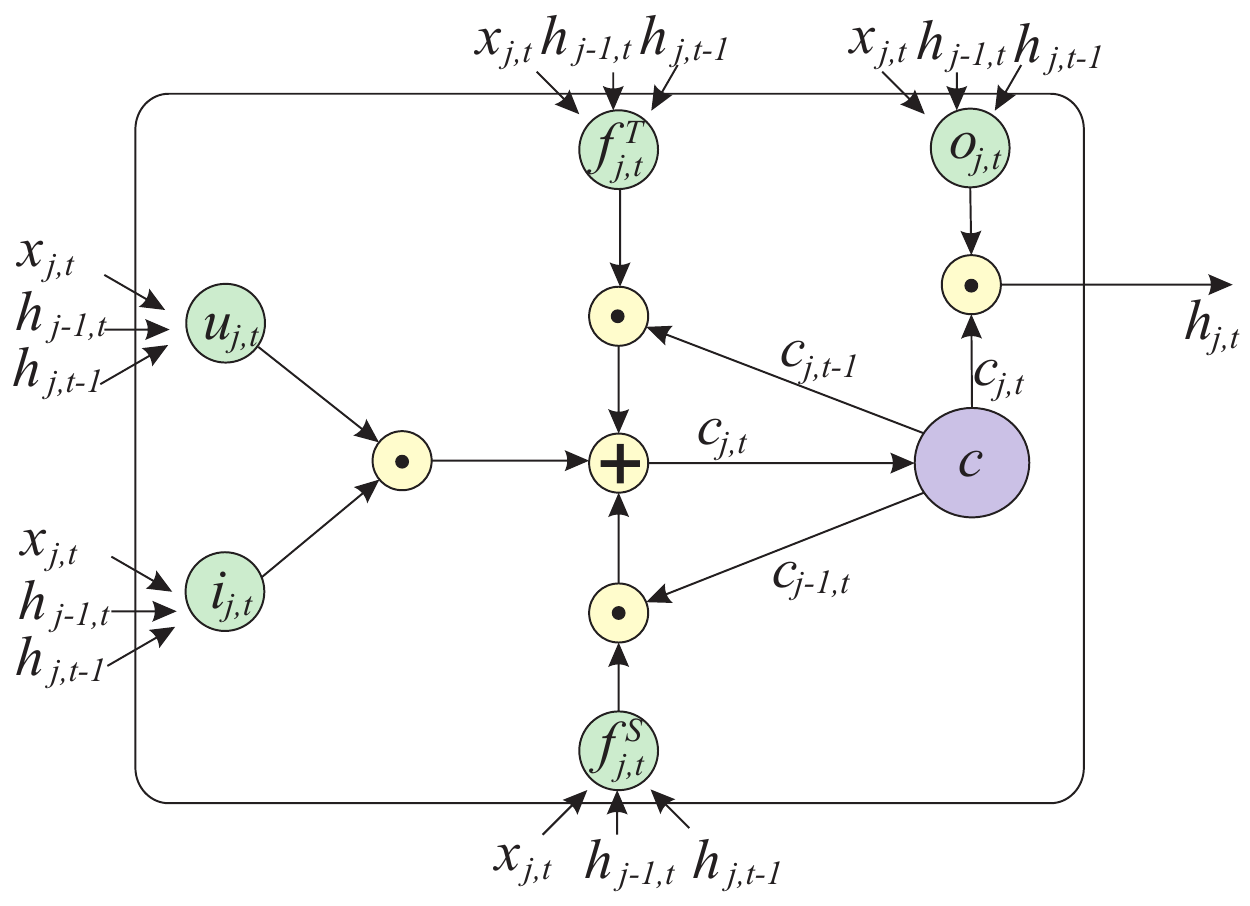}}
	%\end{minipage}
	%\begin{minipage}[b]{0.5\linewidth}
	%  \centering
	%  \centerline{\includegraphics[scale=0.42]{TrustGateSTLSTMFig_U.pdf}}
	%\end{minipage}
	\caption{Illustration of the proposed ST-LSTM with one unit.}
	\label{fig:STLSTMFig}
\end{figure}

As depicted in \figurename{ \ref{fig:STLSTMFig}},
each unit also has two forget gates, $f_{j, t}^{T}$ and $f_{j, t}^{S}$, to handle the two sources of context information in temporal and spatial dimensions, respectively.
%$f_{j, t}^{T}$ for the temporal dimension and $f_{j, t}^{S}$ for the spatial dimension.
%In the spatial dimension, the joints of the skeleton are fed to the network one by one, i.e., at each location $(j, t)$, the coordinates of the joint $j$ in frame $t$ are used as $x_{j,t}$ input.
%The operations performed by spatial-temporal LSTM neuron can be formulated as:
The transition equations of ST-LSTM are formulated as follows:
\begin{eqnarray}
\left(
   \begin{array}{ccc}
    i_{j, t} \\
    f_{j, t}^{S} \\
    f_{j, t}^{T} \\
    o_{j, t} \\
    u_{j, t} \\
   \end{array}
\right)
&=&
\left(
   \begin{array}{ccc}
    \sigma \\
    \sigma \\
    \sigma \\
    \sigma \\
    \tanh \\
   \end{array}
\right)
\left(
   M
   \left(
       \begin{array}{ccc}
        x_{j, t} \\
        h_{j-1, t} \\
        h_{j, t-1} \\
       \end{array}
   \right)
\right)
\\
c_{j, t} &=&  i_{j, t} \odot u_{j, t} + f_{j, t}^{S} \odot  c_{j-1, t} + f_{j, t}^{T} \odot  c_{j, t-1}
\\
h_{j, t} &=& o_{j, t}  \odot \tanh( c_{j, t})
\end{eqnarray}

%In contrast to the standard LSTM model defined over time, the memory unit in ST-LSTM has two preceding states $c_{j-1,t}$ (the previous joint at the same time step) and $c_{j,t-1}$ (the same joint at previous time step). Correspondingly, there are two forget gates $f_{j, t}^{S}$ and $f_{j, t}^{T}$, used for controlling the memory over spatial and time domains respectively.

%\textbf{\emph{Professor suggests to delete the following section}} In our spatial-temporal framework, we do not go through all frames of a video over time. For each sequence, we split it equally to $T$ segments, and in each segment we randomly pick one frame to feed it to the network, so the time step number in our network is $T$. The purpose of the random selection is that for each video sample, we can have a large amount of selection combinations for the $T$ frames from the $T$ segments, which is a promising data augmentation strategy for solving over-fitting issue.

\subsection{Tree-Structure Based Traversal}
\label{sec:approach:skeltree}

\begin{figure}
	\begin{minipage}[b]{0.32\linewidth}
		\centering
		\centerline{\includegraphics[scale=.27]{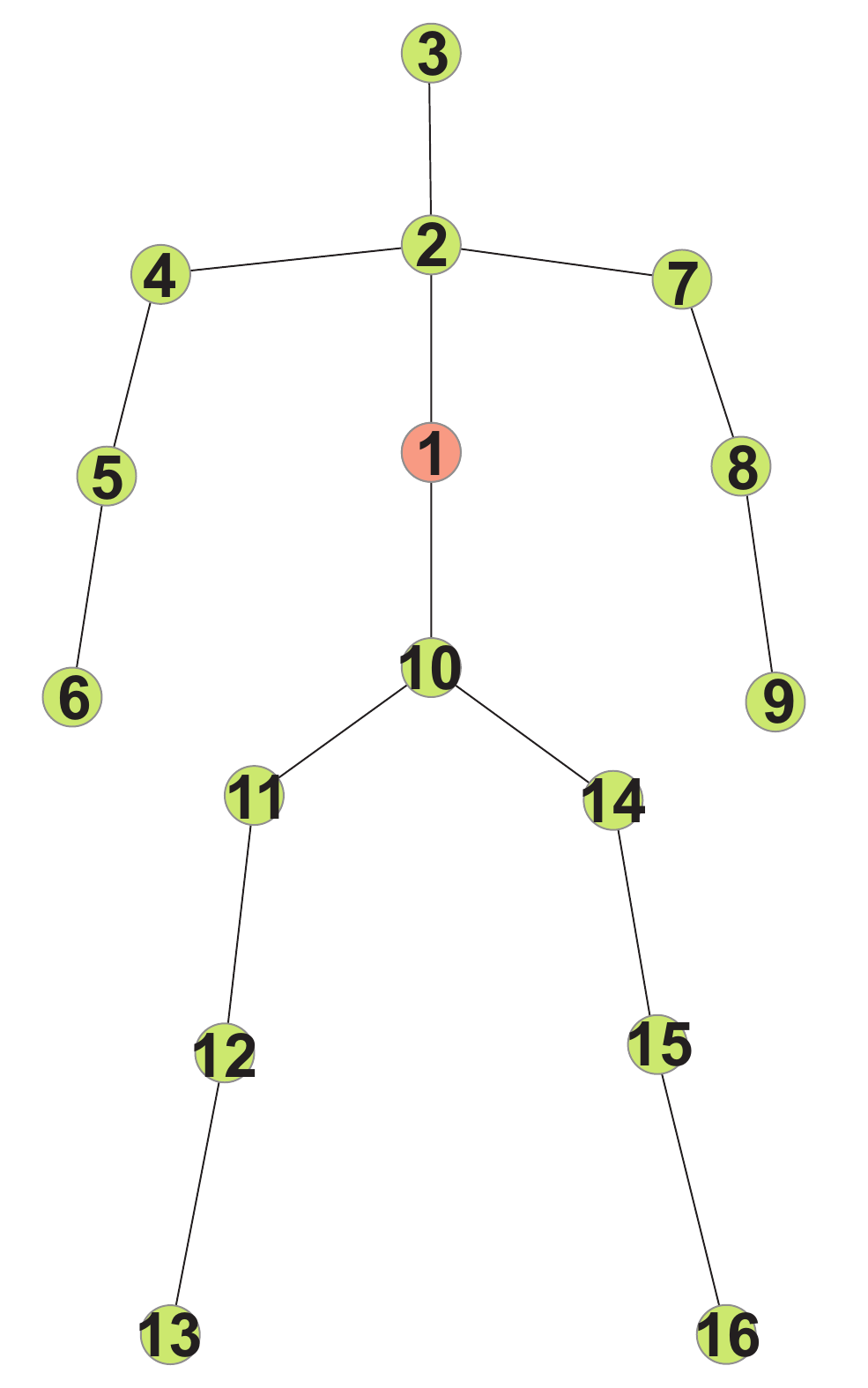}}
		\centerline{(a)}
	\end{minipage}
	\begin{minipage}[b]{0.63\linewidth}
		\centering
		\centerline{\includegraphics[scale=.27]{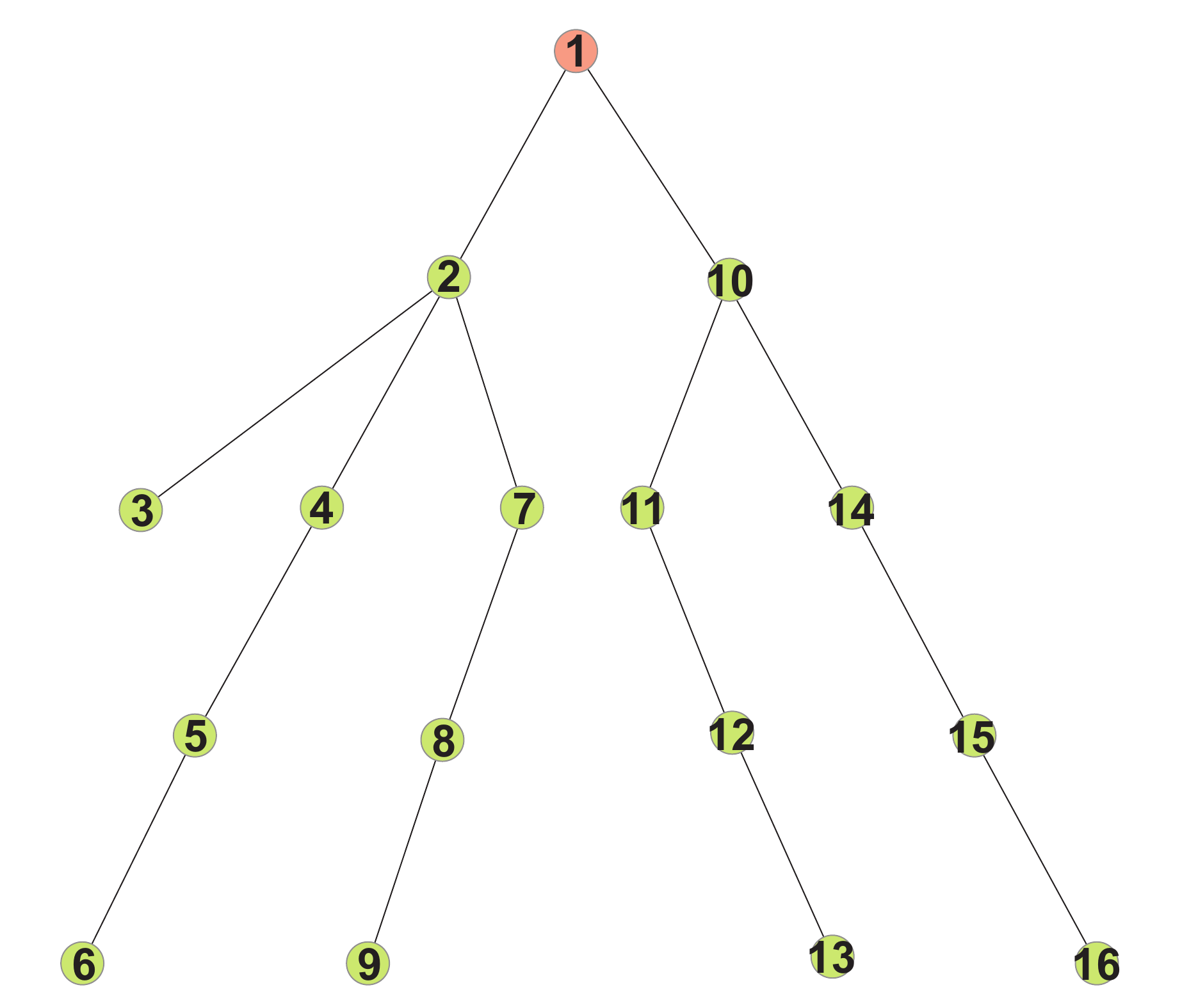}}
		\centerline{(b)}
	\end{minipage}
	\begin{minipage}[b]{0.99\linewidth}
		\centering
		\centerline{\includegraphics[scale=.27]{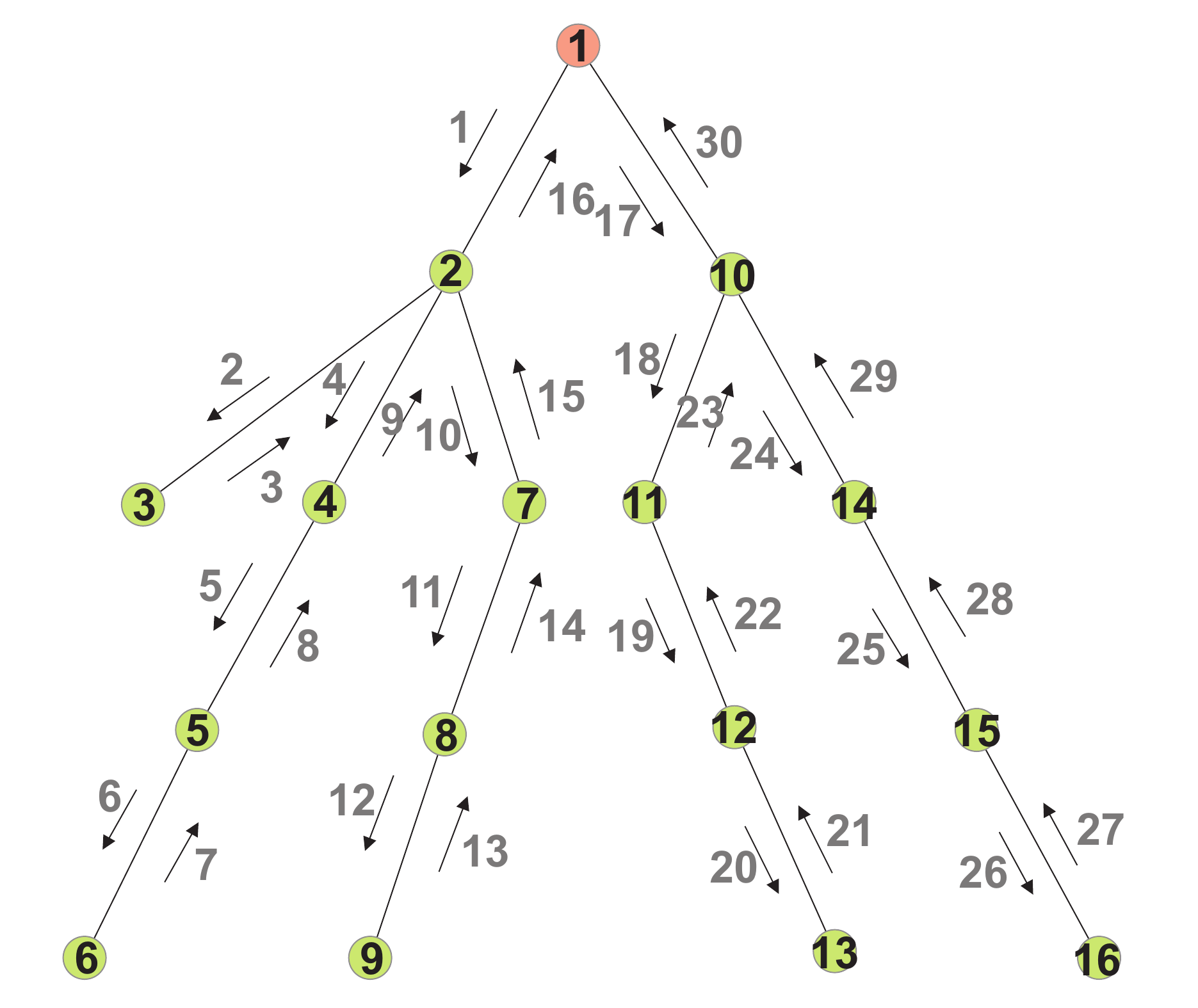}}
		\centerline{(c)}
	\end{minipage}
	\caption{(a) The skeleton of the human body. In the simple joint chain model, the joint visiting order is 1-2-3-...-16.
(b) The skeleton is transformed to a tree structure.
(c) The tree traversal scheme. The tree structure can be unfolded to a chain with the traversal scheme, and the joint visiting order is 1-2-3-2-4-5-6-5-4-2-7-8-9-8-7-2-1-10-11-12-13-12-11-10-14-15-16-15-14-10-1.}
	\label{fig:tree16joints}
\end{figure}

Arranging the skeletal joints in a simple chain order ignores the kinematic interdependencies among the body joints.
Moreover, several semantically false connections between the joints, which are not strongly related, are added.

The body joints are popularly represented as a tree-based pictorial structure \cite{zou2009automatic,yang2011articulated} in human parsing,
as shown in \figurename{ \ref{fig:tree16joints}(b)}.
It is beneficial to utilize the known interdependency relations between various sets of body joints as an adjacency tree structure inside our ST-LSTM network as well.
For instance, the hidden representation of the neck joint (joint 2 in \figurename{ \ref{fig:tree16joints}(a)})
is often more informative for the right hand joints (7, 8, and 9) compared to the joint 6, which lies before them in the numerically ordered chain-like model.
Although using a tree structure for the skeletal data sounds more reasonable here, tree structures cannot be directly fed into our current form of the proposed ST-LSTM network.

In order to mitigate the aforementioned limitation, a bidirectional tree traversal scheme is proposed.
In this scheme, the joints are visited in a sequence, while the adjacency information in the skeletal tree structure will be maintained.
At the first spatial step, the root node (central spine joint in \figurename{ \ref{fig:tree16joints}(c)}) is fed to our network.
Then the network follows the depth-first traversal order in the spatial (skeleton tree) domain. % strategy.
Upon reaching a leaf node, the traversal backtracks in the tree.
Finally, the traversal goes back to the root node.
%By using this traversal scheme, each connection in the tree is passed twice, and the context information is transmitted along both directions.
%when a node is visited for the second time, the network already contains the context representation from father node and child node.
%Upon the end of the traversal, it gets back to the root node.
%The holistic information of the skeleton will corporate in the root node.
%This traversal strategy enables the ST-LSTM to discover better long-range patterns along spatial direction. %learn the spatial structure and joint dependence of the body. Delete this sentence?}
%The intuition behind this traversal method is similar to that of bi-directional LSTM, in which both history and future information are combined in each time step.
%It is worth noting that when the state is passed from a joint to another joint in the tree, meanwhile, it also needs to be passed to the next time step in our spatial-temporal LSTM model.
%Our tree traversal is easy to implement. Although it models the tree structure of the skeleton, the complicated tree structure can be unfolded to a chain, in which some joints are input at more than one spatial step.

In our traversal scheme, each connection in the tree is met twice,
thus it guarantees the transmission of the context data in both top-down and bottom-up directions within the adjacency tree structure.
In other words, each node (joint) can obtain the context information from both its ancestors and descendants in the hierarchy defined by the tree structure.
Compared to the simple joint chain order described in section \ref{sec:approach:stlstm},
this tree traversal strategy, which takes advantage of the joints' adjacency structure, can discover stronger long-term spatial dependency patterns in the skeleton sequence.

\begin{figure}
	\begin{minipage}[b]{0.99\linewidth}
		\centering
		\centerline{\includegraphics[scale=.38]{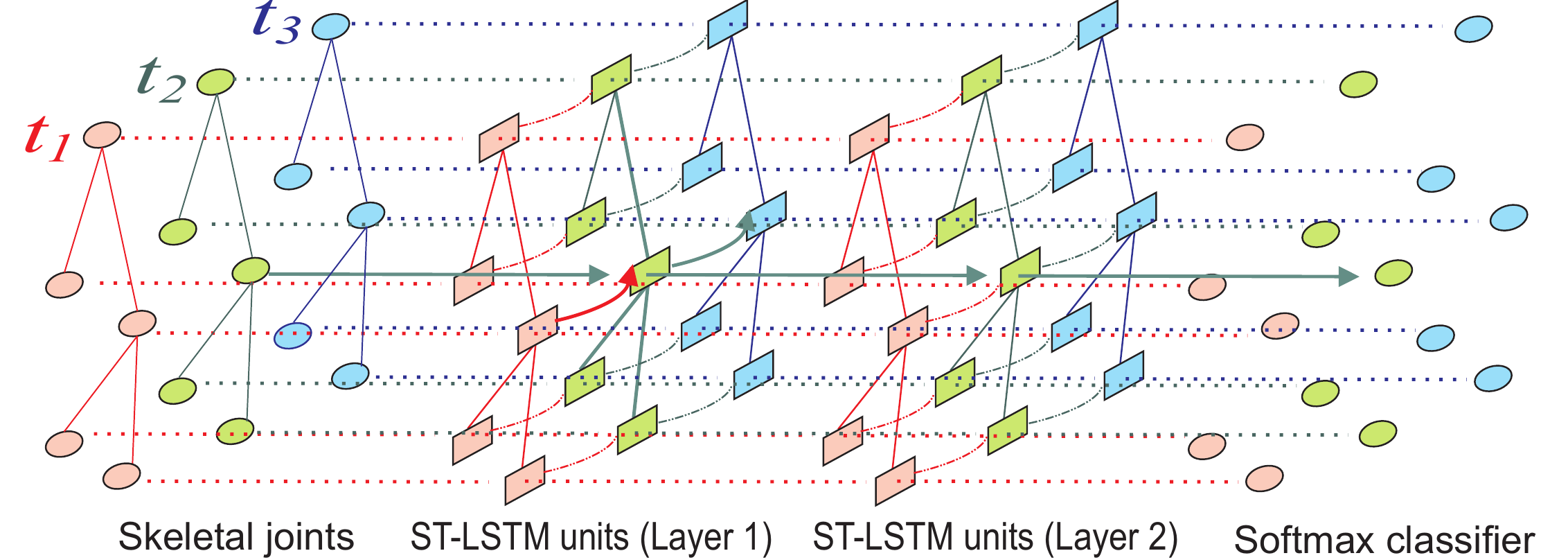}}
	\end{minipage}
	\caption{
Illustration of the deep tree-structured ST-LSTM network.
For clarity, some arrows are omitted in this figure.
The hidden representation of the first ST-LSTM layer is fed to the second ST-LSTM layer as its input.
The second ST-LSTM layer's hidden representation is fed to the softmax layer for classification.
%The overall classification is performed by averaging the classification scores of all the steps.
}
	\label{fig:stackedTreeSTLSTM}
\end{figure}

%Similar to other LSTM implementations \cite{graves2013speechICASSP,sutskever2014sequence},
Our framework's representation capacity can be further improved by stacking multiple layers of the tree-structured ST-LSTMs and making the network deeper, as shown in \figurename{ \ref{fig:stackedTreeSTLSTM}}.

It is worth noting that at each step of our ST-LSTM framework,
the input is limited to the information of a single joint at a time step,
and its dimension is much smaller compared to the concatenated input features used by other existing methods.
Therefore, our network has much fewer learning parameters.
This can be regarded as a weight sharing regularization for our learning model,
which leads to better generalization in the scenarios with relatively small sets of training samples.
This is an important advantage for skeleton-based action recognition, since the numbers of training samples in most existing datasets are limited.

\subsection{Spatio-Temporal LSTM with Trust Gates}
\label{sec:approach:trustgate}

%The current structure of network is sensitive to noisy skeletons.
%Skeletons are always noisy because they are calculated by Kinect on real-time and specifically for gaming purposes.
%This affects the performance of our action recognition framework, because ....
%
%To remedy this issue, we propose to add a new gate to the traditional LSTM unit...
%Talk about trust gate...
%
%Give the specific and technical details; equations; figures; ...

In our proposed tree-structured ST-LSTM network, the inputs are the positions of body joints provided by depth sensors (such as Kinect),
which are not always accurate because of noisy measurements and occlusion.
The unreliable inputs can degrade the performance of the network.

To circumvent this difficulty, we propose to add a novel additional gate to our ST-LSTM network to analyze the reliability of the input measurements based on the derived estimations of the input from the available context information at each spatio-temporal step.
%Our methodology is to make the network capable of predicting the next input, and then measure the difference between the real input and the prediction. If the difference is small, we can consider that the new input is trustable. Oppositely, if the new input has very large difference to the prediction, the network will try to use the trust gate to limit the effect of the unexpected input (probably noisy input) on the network.
Our gating scheme is inspired by the works in natural language processing \cite{sutskever2014sequence},
which use the LSTM representation of previous words at each step to predict the next coming word.
As there are often high dependency relations among the words in a sentence, this idea works decently.
Similarly, in a skeletal sequence, the neighboring body joints often move together,
and this articulated motion follows common yet complex patterns,
thus the input data $x_{j,t}$ is expected to be predictable by using the contextual information ($h_{j-1,t}$ and $h_{j,t-1}$) at each spatio-temporal step.

Inspired by this predictability concept, we add a new mechanism to our ST-LSTM calculating a prediction of the input at each step and comparing it with the actual input.
The amount of estimation error is then used to learn a new ``trust gate''.
The activation of this new gate can be used to assist the ST-LSTM network to learn better decisions about when and how to remember or forget the contents in the memory cell.
For instance, if the trust gate learns that the current joint has wrong measurements,
then this gate can block the input gate and prevent the memory cell from being altered by the current unreliable input data.
%The prediction of an input in a frame can be formulated as:

Concretely, we introduce a function to produce a prediction of the input at step $(j,t)$ based on the available context information as:
\begin{equation}
p_{j, t} = \tanh
\left(
   M_{p}
   \left(
       \begin{array}{ccc}
        h_{j-1, t} \\
        h_{j, t-1} \\
       \end{array}
   \right)
\right)
\label{eq:p_j_t}
\end{equation}
where $M_p$ is an affine transformation mapping the data from $\mathbb{R}^{2d}$ to $\mathbb{R}^d$, thus the dimension of $p_{j,t}$ is $d$.
%It is worth noting that as the prediction is not only based on the hidden states from previous spatial step but also from previous temporal step, i.e., the coordinates of the same joint in previous frames and the other joints visited already in the same frame are seamlessly incorporated, so the network can produce reliable prediction.
Note that the context information at each step does not only contain the representation of the previous temporal step,
but also the hidden state of the previous spatial step.
This indicates that the long-term context information of both the same joint at previous frames and the other visited joints at the current frame are seamlessly incorporated.
Thus this function is expected to be capable of generating reasonable predictions.

In our proposed network, the activation of trust gate is a vector in $\mathbb{R}^d$ (similar to the activation of input gate and forget gate).
The trust gate $\tau_{j, t}$ is calculated as follows:
\begin{eqnarray}
x'_{j, t} &=& \tanh
\left(
   M_{x}
   \left(
        x_{j, t}
   \right)
\right)
\label{eq:x_prime_j_t}
\\
\tau_{j, t} &=& G (p_{j, t} - x'_{j, t})
\label{eq:tau}
%\tau_{j, t} =  G (u_{j, t} - p_{j, t})
%\label{eq:tau}
\end{eqnarray}
%In this equation, $u_{j, t}$ is transformed from the history hidden states and the new input, while $p_{j, t}$ is just from history hidden states, so we can use $(u_{j, t} - p_{j, t})$ to measure the difference between the prediction and real input.
where $M_x: \mathbb{R}^{D} \to \mathbb{R}^{d}$ is an affine transformation.
The activation function $G(\cdot)$ is an element-wise operation calculated as $G(z) = \exp(-\lambda z^{2})$,
for which $\lambda$ is a parameter to control the bandwidth of Gaussian function ($\lambda > 0$).
$G(z)$ produces a small response if $z$ has a large absolute value and a large response when $z$ is close to zero.

Adding the proposed trust gate, the cell state of ST-LSTM will be updated as:
\begin{eqnarray}
c_{j, t} &=&  \tau_{j, t} \odot i_{j, t} \odot u_{j, t}
\nonumber\\
         &&+ (\bold{1} - \tau_{j, t}) \odot f_{j, t}^{S} \odot  c_{j-1, t}
\nonumber\\
         &&+ (\bold{1} - \tau_{j, t}) \odot f_{j, t}^{T} \odot  c_{j, t-1}
\end{eqnarray}

This equation can be explained as follows:
(1) if the input $x_{j,t}$ is not trusted (due to the noise or occlusion),
then our network relies more on its history information, and tries to block the new input at this step;
(2) on the contrary, if the input is reliable, then our learning algorithm updates the memory cell regarding the input data.
%In traditional LSTM, the effect of the input and history states on the neuron is controlled by the input gate and forget gate, however in our design, it will also be controlled by the third gate - trust gate. As this gate is specifically designed for blocking the noisy input, it will be more suitable for dealing with noisy samples.

\begin{figure}
	%\begin{minipage}[b]{0.5\linewidth}
	% \centering
	\centerline{\includegraphics[scale=0.479]{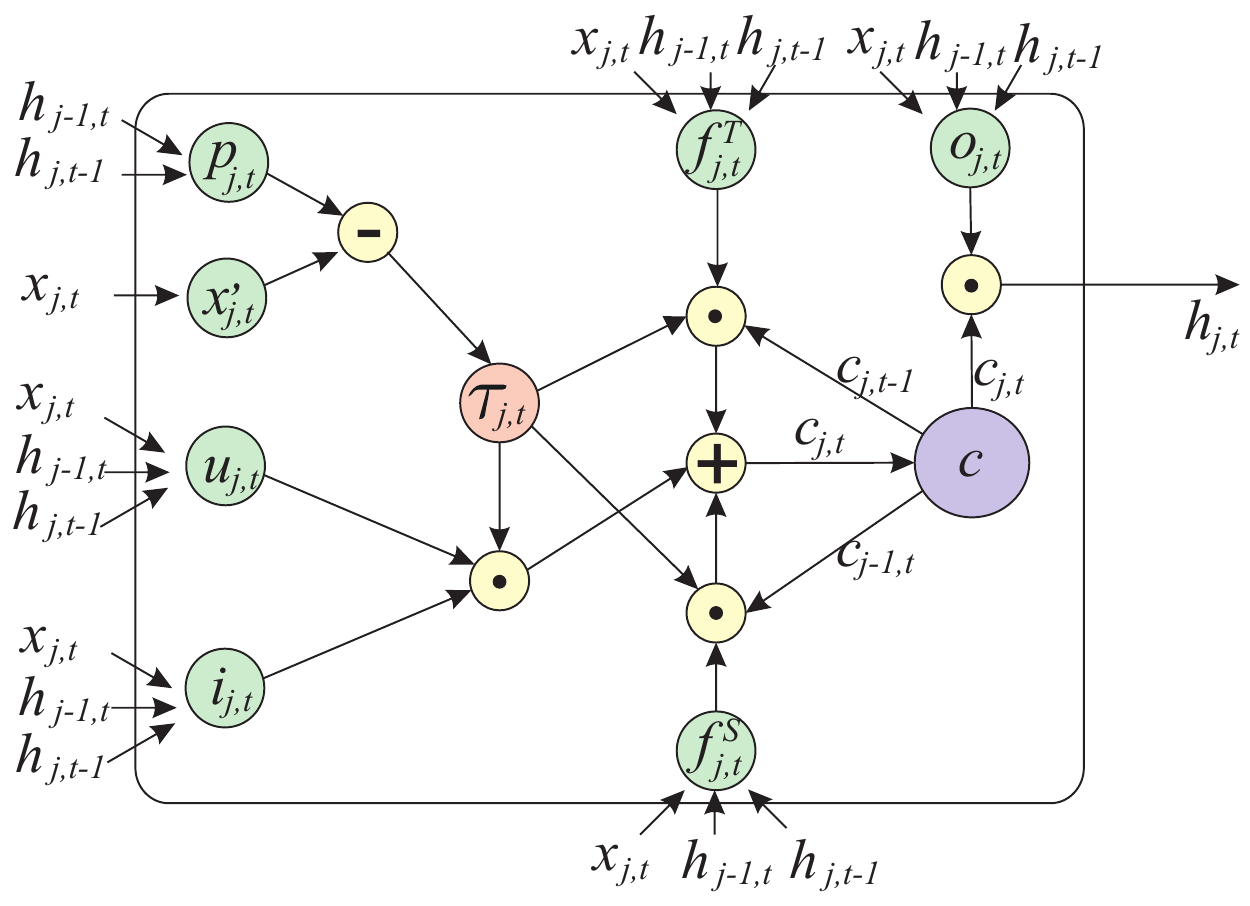}}
	%\end{minipage}
	%\begin{minipage}[b]{0.5\linewidth}
	%  \centering
	%  \centerline{\includegraphics[scale=0.42]{TrustGateSTLSTMFig_U.pdf}}
	%\end{minipage}
	\caption{Illustration of the proposed ST-LSTM with trust gate.}
	\label{fig:TrustGateSTLSTMFig}
\end{figure}

%This new gate (\figurename{ \ref{fig:TrustGateSTLSTMFig}}) can help to deal with noisy samples and can be applied to other applications.
The proposed ST-LSTM unit equipped with trust gate is illustrated in \figurename{ \ref{fig:TrustGateSTLSTMFig}}.
The concept of the proposed trust gate technique is theoretically generic and can be used in other domains to handle noisy input information for recurrent network models.

\subsection{Feature Fusion within ST-LSTM Unit}
\label{sec:approach:innerfusion}

\begin{figure}
	\centerline{\includegraphics[scale=0.469]{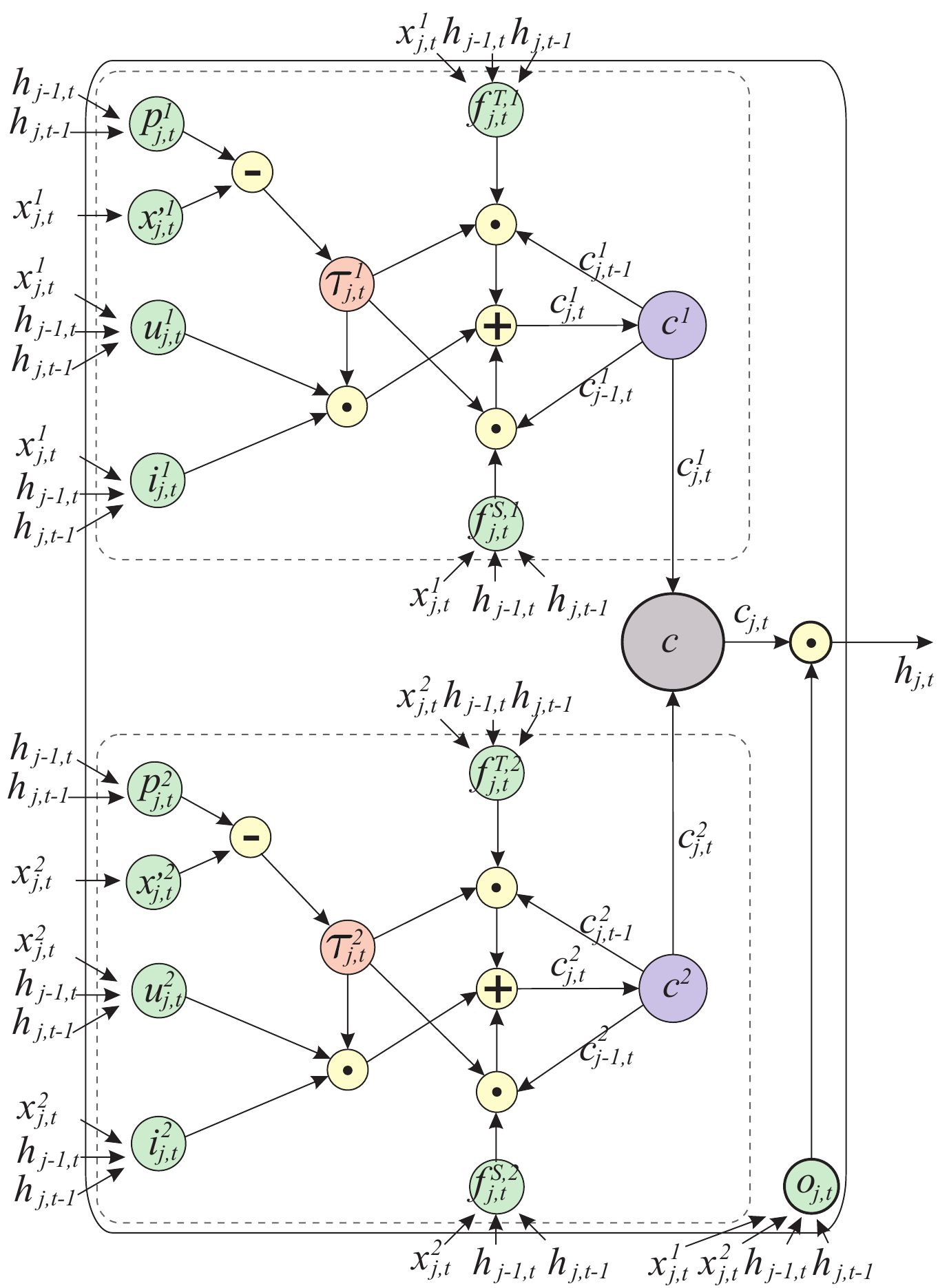}}
	\caption{Illustration of the proposed structure for feature fusion inside the ST-LSTM unit.}
	\label{fig:FusionSTLSTMFig}
\end{figure}

As mentioned above, at each spatio-temporal step, the positional information of the corresponding joint at the current frame is fed to our ST-LSTM network.
Here we call joint position-based feature as a geometric feature.
Beside utilizing the joint position (3D coordinates),
we can also extract visual texture and motion features (\eg HOG, HOF \cite{dalal2006human,wang2011action}, or ConvNet-based features \cite{simonyan2014very,cheron2015p})
from the RGB frames, around each body joint as the complementary information.
This is intuitively effective for better human action representation, especially in the human-object interaction scenarios.

%In existing pose-based approaches,
%the coordinates of the joints \cite{du2015hierarchical, zhu2016co} (or descriptors \cite{lillo2014discriminative} calculated from the joints' coordinates) are often used as the feature (we call them as geometric features).
%Some works \cite{lillo2016hierarchical,} also suggest to extract visual features (such as HOG, HOF, or CNN features) around each joint, and combine them with the geometric features to deal with fine-grained sceneries.
%The two sources of features are often simply concatenated for classification.

A naive way for combining geometric and visual features for each joint is to concatenate them in the feature level
%For our proposed ST-LSTM network, it is straightforward to concatenate the geometric and visual features for each joint
and feed them to the corresponding ST-LSTM unit as network's input data.
However, the dimension of the geometric feature is very low intrinsically,
while the visual features are often in relatively higher dimensions.
Due to this inconsistency, simple concatenation of these two types of features in the input stage of the network causes degradation in the final performance of the entire model.

The work in \cite{nturgbd} feeds different body parts into the Part-aware LSTM \cite{nturgbd} separately,
and then assembles them inside the LSTM unit.
Inspired by this work, we propose to fuse the two types of features inside the ST-LSTM unit,
rather than simply concatenating them at the input level.

%\emph{(Need to talk about some biological inspiration? Different brain areas have different functions: sound, vision, etc. The final resultant information will be also collected and processed by a common area?)}
%% I don't think so! It will not be interesting to talk about this here! And we are not experts in cognitive sciences, our claim can be disastrously wrong!

We use $x_{j,t}^{\mathcal{F}}$ (${\mathcal{F}} \in \{1,2\}$) to denote the geometric feature and visual feature for a joint at the $t$-th time step.
As illustrated in \figurename{ \ref{fig:FusionSTLSTMFig}}, at step $(j,t)$, the two features $(x_{j,t}^{1}$ and $x_{j,t}^{2})$ are fed to the ST-LSTM unit separately as the new input structure.
Inside the recurrent unit, we deploy two sets of gates, input gates $(i_{j,t}^{\mathcal{F}})$, forget gates with respect to time $(f_{j,t}^{T, \mathcal{F}})$ and space $(f_{j,t}^{S, \mathcal{F}})$, and also trust gates $(\tau_{j, t}^{\mathcal{F}})$, to deal with the two heterogeneous sets of modality features.
We put the two cell representations $(c_{j,t}^{\mathcal{F}})$ together to build up the multimodal context information of the two sets of modality features.
Finally, the output of each ST-LSTM unit is calculated based on the multimodal context representations,
and controlled by the output gate $(o_{j,t})$ which is shared for the two sets of features.

For the features of each modality, it is efficient and intuitive to model their context information independently.
However, we argue that the representation ability of each modality-based sets of features can be strengthened by borrowing information from the other set of features.
Thus, the proposed structure does not completely separate the modeling of multimodal features.

Let us take the geometric feature as an example.
Its input gate, forget gates, and trust gate are all calculated from the new input $(x_{j,t}^{1})$ and hidden representations $(h_{j,t-1}$ and $h_{j-1,t})$,
whereas each hidden representation is an associate representation of two features' context information from previous steps.
Assisted by visual features' context information,
the input gate, forget gates, and also trust gate for geometric feature can effectively learn how to update its current cell state $(c_{j,t}^{1})$.
Specifically, for the new geometric feature input $(x_{j,t}^{1})$,
we expect the network to produce a better prediction when it is not only based on the context of the geometric features, but also assisted by the context of visual features.
Therefore, the trust gate $(\tau_{j, t}^{1})$ will have stronger ability to assess the reliability of the new input data $(x_{j,t}^{1})$.

The proposed ST-LSTM with integrated multimodal feature fusion is formulated as:

\begin{eqnarray}
\left(
   \begin{array}{ccc}
    i_{j, t}^\mathcal{F} \\
    f_{j, t}^{S,\mathcal{F}} \\
    f_{j, t}^{T,\mathcal{F}} \\
    u_{j, t}^\mathcal{F} \\
   \end{array}
\right)
&=&
\left(
   \begin{array}{ccc}
    \sigma \\
    \sigma \\
    \sigma \\
    \tanh \\
   \end{array}
\right)
\left(
   M^\mathcal{F}
   \left(
       \begin{array}{ccc}
        x_{j, t}^\mathcal{F} \\
        h_{j-1, t} \\
        h_{j, t-1} \\
       \end{array}
   \right)
\right)
\\
p_{j, t}^\mathcal{F} &=& \tanh
\left(
   M_{p}^\mathcal{F}
   \left(
       \begin{array}{ccc}
        h_{j-1, t} \\
        h_{j, t-1} \\
       \end{array}
   \right)
\right)
\\
{x'}_{j, t}^\mathcal{F} &=& \tanh
\left(
   M_{x}^\mathcal{F}
   \left(
       \begin{array}{ccc}
        x_{j, t}^\mathcal{F}\\
       \end{array}
   \right)
\right)
\\
\tau_{j, t}^{\mathcal{F}} &=& G ({x'}_{j, t}^{\mathcal{F}} - p_{j, t}^{\mathcal{F}})
\\
c_{j, t}^{\mathcal{F}} &=&  \tau_{j, t}^{\mathcal{F}} \odot i_{j, t}^{\mathcal{F}} \odot u_{j, t}^{\mathcal{F}}
\nonumber\\
         &&+ (\bold{1} - \tau_{j, t}^{\mathcal{F}}) \odot f_{j, t}^{S,\mathcal{F}} \odot  c_{j-1, t}^{\mathcal{F}}
\nonumber\\
         &&+ (\bold{1} - \tau_{j, t}^{\mathcal{F}}) \odot f_{j, t}^{T,\mathcal{F}} \odot  c_{j, t-1}^{\mathcal{F}}
\\
o_{j, t} &=& \sigma
\left(
M_{o}
   \left(
   \begin{array}{ccc}
        x_{j, t}^{1} \\
        x_{j, t}^{2} \\
        h_{j-1, t} \\
        h_{j, t-1} \\
   \end{array}
   \right)
\right)
\\
h_{j, t} &=& o_{j, t}  \odot \tanh
   \left(
       \begin{array}{ccc}
        c_{j, t}^{1} \\
        c_{j, t}^{2} \\
       \end{array}
   \right)
\end{eqnarray}

\subsection{Learning the Classifier}
\label{sec:approach:learning}
As the labels are given at video level, we feed them as the training outputs of our network at each spatio-temporal step.
A softmax layer is used by the network to predict the action class $\hat{y}$ among the given class set $Y$.
The prediction of the whole video can be obtained by averaging the prediction scores of all steps.
%Empirically, this method provides better performance compared to the minimization of the loss at the last step only.
The objective function of our ST-LSTM network is as follows:
\begin{equation}
\mathcal{L} = \sum_{j=1}^J \sum_{t=1}^T l(\hat{y}_{j,t}, y)
\end{equation}
where $l(\hat{y}_{j,t}, y)$ is the negative log-likelihood loss \cite{graves2012supervised}
that measures the difference between the prediction result $\hat{y}_{j,t}$ at step $(j,t)$ and the true label $y$.
%The objective function can be minimized using back-propagation through time (BPTT) algorithm \cite{graves2012supervised}.

The back-propagation through time (BPTT) algorithm \cite{graves2012supervised} is often effective for minimizing the objective function for the RNN/LSTM models.
As our ST-LSTM model involves both spatial and temporal steps, we adopt a modified version of BPTT for training.
The back-propagation runs over spatial and temporal steps simultaneously by starting at the last joint at the last frame.
To clarify the error accumulation in this procedure, we use $e_{j,t}^T$ and $e_{j,t}^S$ to denote the error back-propagated from step $(j,t+1)$ to $(j,t)$ and the error back-propagated from step $(j+1,t)$ to $(j,t)$, respectively.
Then the errors accumulated at step $(j,t)$ can be calculated as $e_{j,t}^T+e_{j,t}^S$.
Consequently, before back-propagating the error at each step, we should guarantee both its subsequent joint step and subsequent time step have already been computed.

%\subsection{Further improvement}
%\label{sec:approach:further}

%To improve the performance of the proposed network, we also employ some other strategies, as follows:

%In action recognition, the pose in each frame and the motion pattern among frames are both crucial for accurate recognition. The aforementioned framework utilizes the pose cue (3D locations of the joints), and also the long-range motion. (As $T$ frames are sampled from a video sequence rather than using all frames, the network learns the long-range motion over the time steps). We also want to use the short-range motion for better recognition performance. To make use of the short-range motion cue, for each spatial step at $t$, we not only input the current joint position, but also two motion vectors for this joint. One is the past motion with regard to frame at $t+\Delta t_1$, and the other is the upcoming motion with regard to frame at $t+\Delta t_2$. In our experiment, we do not fix $\Delta t_1$ or $\Delta t_2$, but use small random values for them considering data augmentation. The motion can be obtained by subtracting the positions of the same joint in each two frames. For efficiency, we directly input the joint positions in the three frames without executing subtraction in our method, and believe the network is able to discover the motion in the frames.

The left-most units in our ST-LSTM network do not have preceding spatial units, as shown in \figurename{ \ref{fig:STLSTM}}.
To update the cell states of these units in the feed-forward stage,
a popular strategy is to input zero values into these nodes to substitute the hidden representations from the preceding nodes.
In our implementation, we link the last unit at the last time step to the first unit at the current time step.
%, as shown in \figurename{ \ref{fig:STLSTM_LinkLast}}.
We call the new connection as last-to-first link.
In the tree traversal, the first and last nodes refer to the same joint (root node of the tree),
however the last node contains holistic information of the human skeleton in the corresponding frame.
Linking the last node to the starting node at the next time step provides the starting node with the whole body structure configuration,
rather than initializing it with less effective zero values.
Thus, the network has better ability to learn the action patterns in the skeleton sequence.

\section{Experiments}
\label{sec:exp}

%Talk about the details and structure of the input data.  Their properties, how they are captured, what is wrong with them...

The proposed method is evaluated and empirically analyzed on seven benchmark datasets for which the coordinates of skeletal joints are provided.
These datasets are NTU RGB+D, UT-Kinect, SBU Interaction, SYSU-3D, ChaLearn Gesture, MSR Action3D, and Berkeley MHAD.
%We additionally evaluate the proposed method on the Sub-JHMDB dataset, in which the 2D joint positions are extracted by using a pose estimation algorithm, to verify the effectiveness of the proposed method for action recognition in RGB videos.
We conduct extensive experiments with different models to verify the effectiveness of individual technical contributions proposed, as follows:
%\begin{enumerate}

(1) ``ST-LSTM (Joint Chain)''.
  In this model, the joints are visited in a simple chain order, as shown in \figurename{ \ref{fig:tree16joints}(a)};

(2) ``ST-LSTM (Tree)''.
  In this model, the tree traversal scheme illustrated in \figurename{ \ref{fig:tree16joints}(c) is used to take advantage of the tree-based spatial structure of skeletal joints;

(3) ``ST-LSTM (Tree) + Trust Gate''.
  This model uses the trust gate to handle the noisy input.
%\end{enumerate}

%We also design the following baselines to further analysis the the performance of our method:

The input to every unit of of our network at each spatio-temporal step is the location of the corresponding skeletal joint (i.e., geometric features) at the current time step.
We also use two of the datasets (NTU RGB+D dataset and UT-Kinect dataset) as examples
to evaluate the performance of our fusion model within the ST-LSTM unit by fusing the geometric and visual features.
These two datasets include human-object interactions (such as making a phone call and picking up something)
and the visual information around the major joints can be complementary to the geometric features for action recognition.

%For the datasets with 3D skeletons, the input of our network at each spatio-temporal step is the 3D coordinates of the corresponding joint (geometric features).

%For the RGB video dataset, we evaluate the performance of the proposed fusion model within ST-LSTM unit for 2D skeletal data (geometric features) and visual features around each body joint in the RGB frames, and compare it with the naive feature level concatenation method.

\subsection{Evaluation Datasets}
\label{sec:exp:datasets}

{\bf NTU RGB+D dataset} \cite{nturgbd} was captured with Kinect (v2).
It is currently the largest publicly available dataset for depth-based action recognition, which contains more than 56,000 video sequences and 4 million video frames.
The samples in this dataset were collected from 80 distinct viewpoints.
A total of 60 action classes (including daily actions, medical conditions, and pair actions) were performed by 40 different persons aged between 10 and 35.
This dataset is very challenging due to the large intra-class and viewpoint variations.
With a large number of samples, this dataset is highly suitable for deep learning based activity analysis.
The parameters learned on this dataset can also be used to initialize the models for smaller datasets to improve and speed up the training process of the network.  %!!!!!!!!!!!!!!!!!!!!!!!!!!!!!!!!!!!!!!
The 3D coordinates of 25 body joints are provided in this dataset.

{\bf UT-Kinect dataset} \cite{HOJ3D} was captured with a stationary Kinect sensor.
It contains 10 action classes. % (walk, sit down, stand up, pick up, carry, throw, push, pull, wave hands, and clap hands) performed by 10 subjects.
Each action was performed twice by every subject.
The 3D locations of 20 skeletal joints are provided.
The significant intra-class and viewpoint variations make this dataset very challenging.

{\bf SBU Interaction dataset} \cite{yun2012two} was collected with Kinect.
It contains 8 classes of two-person interactions, and includes 282 skeleton sequences with 6822 frames.
Each body skeleton consists of 15 joints.
The major challenges of this dataset are:
(1) in most interactions, one subject is acting, while the other subject is reacting; and
(2) the 3D measurement accuracies of the joint coordinates are low in many sequences.

%{\bf CMU Motion Capture Dataset}. This dataset containing 2235 sequences was categorized into 45 classes by \cite{zhu2016co}. Each frame contains 31 skeleton joints. The challenges of this dataset include (1) high complexity of some actions, such as dancing; (2) large intra-class variations; (3) high diversity of sequence lengths. Several different evaluation protocols have been used on CMU dataset by existing work.

{\bf SYSU-3D dataset} \cite{jianfang_CVPR15} contains 480 sequences and was collected with Kinect.
In this dataset, 12 different activities were performed by 40 persons.
The 3D coordinates of 20 joints are provided in this dataset.
The SYSU-3D dataset is a very challenging benchmark because:
(1) the motion patterns are highly similar among different activities, and
(2) there are various viewpoints in this dataset.

{\bf ChaLearn Gesture dataset} \cite{escalera2013multi} consists of 23 hours of videos captured with Kinect.
A total of 20 Italian gestures were performed by 27 different subjects.
%This dataset contains 6850 training samples, 3454 validation samples, and 3579 testing samples.
This dataset contains 955 long-duration videos and has predefined splits of samples as training, validation and testing sets.
Each skeleton in this dataset has 20 joints.

{\bf MSR Action3D dataset} \cite{li2010action} is widely used for depth-based action recognition.
It contains a total of 10 subjects and 20 actions.
%(high arm wave, horizontal arm wave, hammer, hand catch, forward punch, high throw, draw x, draw tick, draw circle, hand clap, two hand wave, side-boxing, bend, forward kick, side kick, jogging, tennis swing, tennis serve, golf swing, and pick up \& throw).
Each action was performed by the same subject two or three times.
Each frame in this dataset contains 20 skeletal joints.

{\bf Berkeley MHAD dataset} \cite{ofli2013berkeley} was collected by using a motion capture network of sensors.
It contains 659 sequences and about 82 minutes of recording time.
Eleven action classes were performed by five female and seven male subjects.
The 3D coordinates of 35 skeletal joints are provided in each frame.

%{\bf Sub-JHMDB} \cite{jhuang2013towards}. This is an RGB video dataset including 12 action classes. In this dataset, most of the actor bodies are fully visible in the image frames. Following \cite{cheron2015p}, we extract the 2D skeletal joints using the pose estimation method in \cite{yang2011articulated}, in which 15 joints are extracted in each RGB frame.

\subsection{Implementation Details}
\label{sec:exp:impdetails}

%%UT-Kinect and NTU datasets are very challenging due to the large amount of variations of view points. To make the skeleton view-invariant, we rotated the skeletons to make the vector from left hip to right hip parallel to the global X-axis for these two datasets. We also performed a translation to the skeleton joints similar to \cite{du2015hierarchical} to make the hip-center located in the origin of the coordinate system. We did not use the aforementioned preprocessing procedures on SBU dataset and MHAD dataset, and the proposed model still obtained state-of-the-art results on these two datasets.
%
%It often helps to get better result if all skeleton joints are fed to the proposed network. However, more input means the network will be larger and need more memory. To make a tradeoff of action recognition accuracy and GPU memory consumption, we omit some joints from the input data and make sure the remaining joints contain enough discriminative information of describing the actions. The joints used in our evaluation are depicted in Fig. \ref{fig:SkeletonAll}.
%
%\begin{figure}
%\begin{minipage}[b]{1.0\linewidth}
%  \centering
%  \centerline{\includegraphics[scale=0.20]{SkeletonAll.pdf}}
%\end{minipage}
%\caption{Skeletons of different datasets. The joints in red and green are used in our experiments. Red indicates the root node of our tree structure. Blue joints are ignored.}
%\label{fig:SkeletonAll}
%\end{figure}

In our experiments, each video sequence is divided to $T$ sub-sequences with the same length, and one frame is randomly selected from each sub-sequence.
This sampling strategy has the following advantages:
(1) Randomly selecting a frame from each sub-sequence can add variation to the input data, and improves the generalization strengths of our trained network.
(2) Assume each sub-sequence contains $n$ frames,
so we have $n$ choices to sample a frame from each sub-sequence.
Accordingly, for the whole video, we can obtain a total number of $n^T$ sampling combinations.
This indicates that the training data can be greatly augmented.
We use different frame sampling combinations for each video over different training epochs.
This strategy is useful for handling the over-fitting issues,
as most datasets have limited numbers of training samples.
We observe this strategy achieves better performance in contrast with uniformly sampling frames.
We cross-validated the performance based on the leave-one-subject-out protocol on the large scale NTU RGB+D dataset, and found $T=20$ as the optimum value.

We use Torch7 \cite{collobert2011torch7} as the deep learning platform to perform our experiments.
We train the network with stochastic gradient descent,
and set the learning rate, momentum, and decay rate to $2$$\times$$10^{-3}$, $0.9$, and $0.95$, respectively.
We set the unit size $d$ to 128, and the parameter $\lambda$ used in $G(\cdot)$ to $0.5$.
Two ST-LSTM layers are used in our stacked network.
Although there are variations in terms of joint number, sequence length, and data acquisition equipment for different datasets,
we adopt the same parameter settings mentioned above for all datasets.
Our method achieves promising results on all the benchmark datasets with these parameter settings untouched, which shows the robustness of our method.

%We evaluate the efficiency of our method on the NTU RGB+D dataset.
%We perform our experiments with an NVIDIA TitanX GPU.
%On the NTU RGB+D dataset,
%each batch can be trained in $3.9s$ with the batch size set to 100.
%Our model ``ST-LSTM (Tree) + Trust Gate'' runs very fast.
%Averagely, each video from NTU RGB+D dataset can be processed by ``ST-LSTM (Tree) + Trust Gate'' model in $1.3s$ on GPU mode ,
%and $3.3s$ on CPU mode.

An NVIDIA TitanX GPU is used to perform our experiments.
We evaluate the computational efficiency of our method on the NTU RGB+D dataset and set the batch size to $100$.
On average, within one second, $210$, $100$, and $70$ videos can be processed
by using ``ST-LSTM (Joint Chain)'', ``ST-LSTM (Tree)'', and ``ST-LSTM (Tree) + Trust Gate'', respectively.

\subsection{Experiments on the NTU RGB+D Dataset}
\label{sec:exp:resNTU}

%Talk about our experiments one by one. Evaluate each of our contributions step by step and show how they improved the results for few number of datasets, preferably including our dataset (since it's in the larger scales and therefore more reliable).

%We perform our experiments on an NVidia Tesla K40 GPU. Using the Torch toolbox, the average running time per frame is 0.08ms, 0.16ms and 0.49ms for ST-LSTM(Joint Chain), ST-LSTM(Tree Traversal), and ST-LSTM(Tree Traversal)+Trust Gate, respectively.

The NTU RGB+D dataset has two standard evaluation protocols \cite{nturgbd}.
The first protocol is the cross-subject (X-Subject) evaluation protocol,
in which half of the subjects are used for training and the remaining subjects are kept for testing.
The second is the cross-view (X-View) evaluation protocol,
in which $2/3$ of the viewpoints are used for training,
and $1/3$ unseen viewpoints are left out for testing.
We evaluate the performance of our method on both of these protocols.
The results are shown in \tablename{ \ref{table:resultNTU}}.

\begin{table}[!htp]
\caption{Experimental results on the NTU RGB+D Dataset}
\label{table:resultNTU}
\centering
%\scriptsize
\begin{tabular}{|l|c|c|c|}
\hline
Method & Feature & X-Subject & X-View  \\
%\hline
\hline
%Super Normal Vector \cite{yang2014super}  &  &  \\
Lie Group \cite{vemulapalli2014liegroup} & Geometric & 50.1\% &  52.8\%  \\
%Skeletal Quads \cite{skeletalQuads} & Geometric & 38.6\%  & 41.4\% \\
Cippitelli \etal \cite{cippitelli2016evaluation} & Geometric & 48.9\%  &  57.7\% \\
Dynamic Skeletons \cite{jianfang_CVPR15}  & Geometric &  60.2\% & 65.2\% \\
FTP \cite{rahmani20163d}  & Geometric &  61.1\% & 72.6\% \\   %  !!!!!!!!!!!!!!!!!! -> FTP
Hierarchical RNN \cite{du2015hierarchical}  & Geometric & 59.1\% & 64.0\%  \\
Deep RNN \cite{nturgbd} & Geometric &  56.3\%  &  64.1\%  \\
Part-aware LSTM \cite{nturgbd} & Geometric & 	62.9\% &	70.3\%  \\
%Deep LSTM \cite{nturgbd} & Geometric &   60.7\% & 67.3\%  \\
\hline
ST-LSTM (Joint Chain)  & Geometric &	61.7\%	& 75.5\% \\
ST-LSTM (Tree) & Geometric & 	65.2\% &	76.1\% \\
ST-LSTM (Tree) + Trust Gate & Geometric & \textbf{69.2\%}	&  \textbf{77.7\%} \\
%%ST-LSTM + Tree + Trust Gate (x) & 	68.7\% & \textbf{77.8\%} \\
%%ST-LSTM + Tree + Trust Gate (u) & 	\textbf{69.2\%}	& 77.\%7 \\
\hline
\end{tabular}
\end{table}

In \tablename{ \ref{table:resultNTU}},
the deep RNN model concatenates the joint features at each frame and then feeds them to the network to model the temporal kinetics, and ignores the spatial dynamics.
As can be seen, both ``ST-LSTM (Joint Chain)'' and ``ST-LSTM (Tree)'' models outperform this method by a notable margin.
It can also be observed that our approach utilizing the trust gate brings significant performance improvement,
because the data provided by Kinect is often noisy and multiple joints are frequently occluded in this dataset.
Note that our proposed models (such as ``ST-LSTM (Tree) + Trust Gate'') reported in this table only use skeletal data as input.

We compare the class specific recognition accuracies of ``ST-LSTM (Tree)'' and ``ST-LSTM (Tree) + Trust Gate'', as shown in \figurename{ \ref{fig:ClassAccuracy_NTU}}.
We observe that ``ST-LSTM (Tree) + Trust Gate'' significantly outperforms ``ST-LSTM (Tree)'' for most of the action classes,
which demonstrates our proposed trust gate can effectively improve the human action recognition accuracy by learning the degrees of reliability over the input data at each time step.

\begin{figure*}
\begin{minipage}[b]{1.0\linewidth}
  \centering
  \centerline{\includegraphics[scale=0.38]{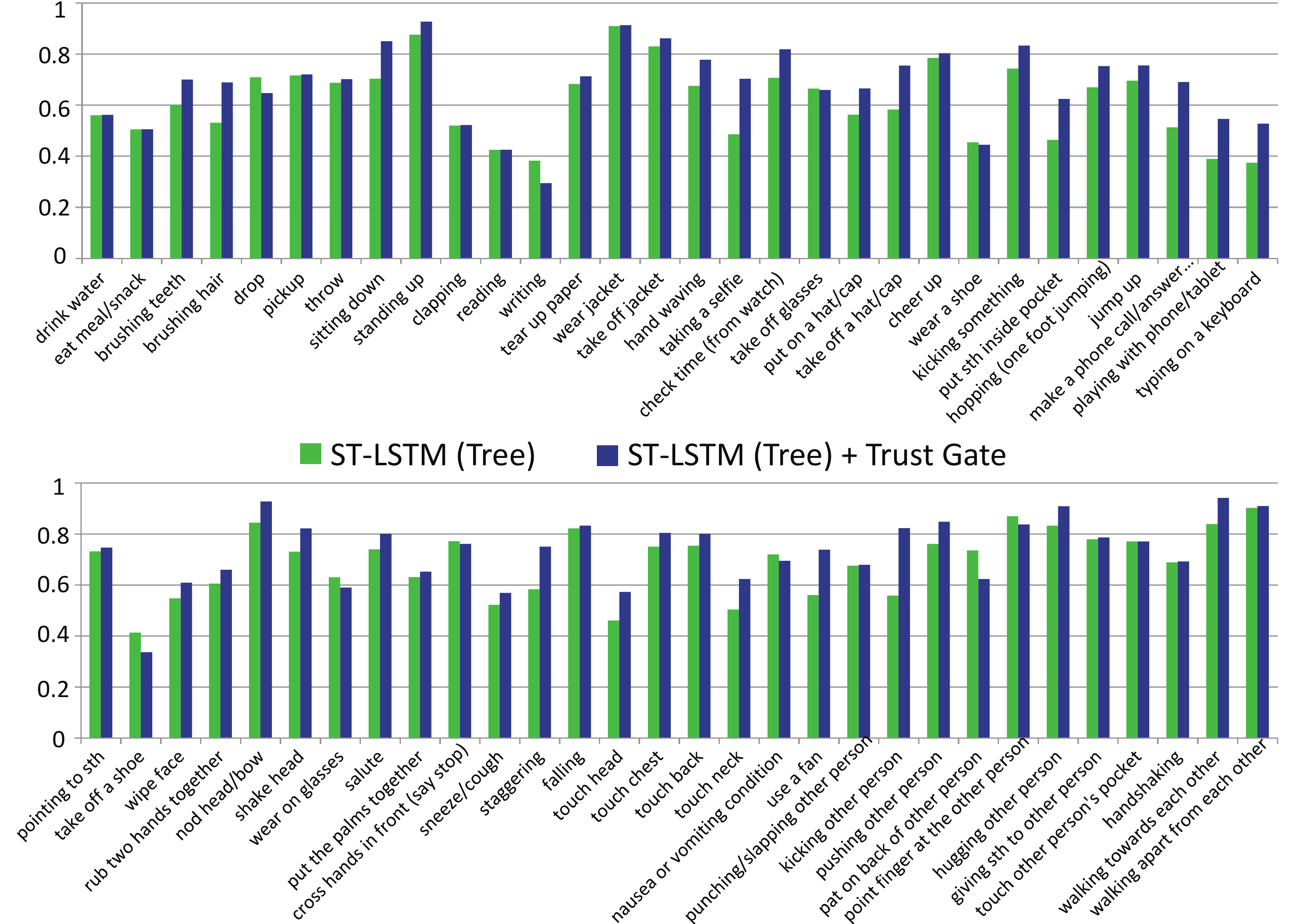}}
\end{minipage}
\caption{Recognition accuracy per class on the NTU RGB+D dataset}
\label{fig:ClassAccuracy_NTU}
\end{figure*}

As shown in \figurename{ \ref{fig:NTUNoisySamples}},
a notable portion of videos in the NTU RGB+D dataset were collected in side views.
Due to the design of Kinect's body tracking mechanism,
skeletal data is less accurate in side view compared to the front view.
To further investigate the effectiveness of the proposed trust gate,
we analyze the performance of the network by feeding the side views samples only.
The accuracy of ``ST-LSTM (Tree)'' is 76.5\%,
while ``ST-LSTM (Tree) + Trust Gate'' yields 81.6\%.
This shows how trust gate can effectively deal with the noise in the input data.

\begin{figure}
\begin{minipage}[b]{1.0\linewidth}
  \centering
  \centerline{\includegraphics[scale=0.199]{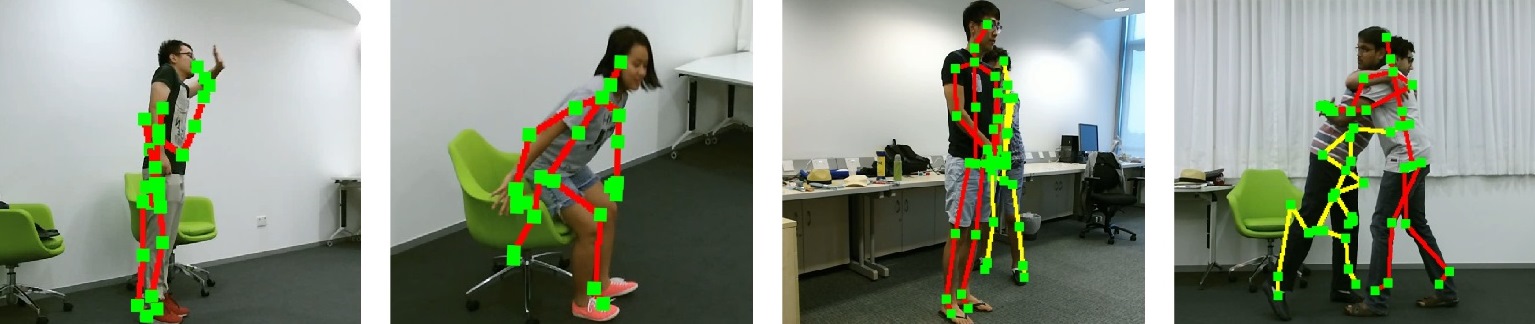}}
\end{minipage}
\caption{Examples of the noisy skeletons from the NTU RGB+D dataset.}
\label{fig:NTUNoisySamples}
\end{figure}

To verify the performance boost by stacking layers,
we limit the depth of the network by using only one ST-LSTM layer,
and the accuracies drop to 65.5\% and 77.0\% based on the cross-subject and cross-view protocol, respectively.
This indicates our two-layer stacked network has better representation power than the single-layer network.

%We also evaluate the performance of the proposed model when different time steps are used. The results are shown in Tab. \ref{table:resultTimeStepNTU}. Using more than 5 time steps, our method can produce very good result. Even if we just use 4 time steps, the accuracy is still quite promising (66.1\%) and is much better than all other methods in Tab. \ref{table:resultNTU}. This indicates that our model does not need too many time steps yet effectively model the temporal dynamics in action sequences. This advantage is partially brought by our short-range motion learning strategy.

%\begin{table}[!hbp]
%\caption{Comparisons of different time steps on NTU dataset (cross subject)}
%\label{table:resultTimeStepNTU}
%\centering
%\begin{tabular}{c|c|c|c|c|c|c|c|c}
%\hline
%Time step      &   4    &  5      &    6  &     7  &      8 &      9  &   10    &       12 \\
%\hline
%Accuracy (\%)  &  66.1  &  67.4   &  69.2 &   68.9 &  68.8  &  69.2   &   69.1  &     69.1  \\
%\hline
%\end{tabular}
%\end{table}

%See Table \ref{table:resultTimeStepNTU}, we can see that 6 time steps are enough for NTU dataset, and this is partially due to in each time step, it includes not only the position of the joint, but also the motion information...  However, performance will drop a bit with too many time steps, this can be explained as: if too many time steps, then when we randomly select frames from a sequence, too many frames, then it will not be random any more, then overfitting may occur.

To evaluate the performance of our feature fusion scheme,
we extract visual features from several regions based on the joint positions and use them in addition to the geometric features (3D coordinates of the joints).
%In this dataset, there are some actions which include human-object interactions, such as catch, pick, and shoot ball,
%thus the visual information around the joints is complementary and discriminative for action classification.
We extract HOG and HOF \cite{dalal2006human,wang2011action} features from a $80\times80$ RGB patch centered at each joint location.
For each joint, this produces a 300D visual descriptor,
and we apply PCA to reduce the dimension to 20.
The results are shown in \tablename{ \ref{table:resultNTUFusion}}.
We observe that our method using the visual features together with the joint positions improves the performance.
Besides, we compare our newly proposed feature fusion strategy within the ST-LSTM unit with two other feature fusion methods:
(1) early fusion which simply concatenates two types of features as the input of the ST-LSTM unit;
(2) late fusion which uses two ST-LSTMs to deal with two types of features respectively,
then concatenates the outputs of the two ST-LSTMs at each step,
and feeds the concatenated result to a softmax classifier.
We observe that our proposed feature fusion strategy is superior to other baselines.

\begin{table}[h]
		\caption{Evaluation of different feature fusion strategies on the NTU RGB+D dataset.
``Geometric + Visual (1)'' indicates the early fusion scheme.
``Geometric + Visual (2)'' indicates the late fusion scheme.
``Geometric $\bigoplus$ Visual'' means our newly proposed feature fusion scheme within the ST-LSTM unit.}
		\label{table:resultNTUFusion}
%\scriptsize
		\centering
		\begin{tabular}{|l|c|c|}
			\hline
			Feature Fusion Method & X-Subject & X-View
              \\
			\hline
			%\hline
			Geometric Only           &  69.2\%	&   77.7\% \\
            Geometric + Visual (1) & 70.8\%   &   78.6\%                  \\
            Geometric + Visual (2) & 71.0\%   &   78.7\%               \\
            Geometric $\bigoplus$ Visual  &73.2\%   &  80.6\%  \\
			\hline
		\end{tabular}
\\
\end{table}

We also evaluate the sensitivity of the proposed network with respect to the variation of neuron unit size and $\lambda$ values.
The results are shown in \figurename{ \ref{fig:NTUResultLambda}}.
When trust gate is added,
our network obtains better performance for all the $\lambda$ values compared to the network without the trust gate.

%\begin{table}[!hbp]
%\caption{Comparisons of different $\lambda$ values on NTU dataset (cross subject)}
%\label{table:resultLambdaNTU}
%\centering
%\begin{tabular}{c|c|c|c|c|c|c|c}
%\hline
%$\lambda$      & $10^{-3}$ & $10^{-2}$  & 0.1       &  0.2     &  0.5   &        1        &    2   \\
%\hline
%Accuracy (\%)  & 66.4      &    66.4    &           &   68.2   &  69.2  &       68.3      &  67.2    \\
%\hline
%\end{tabular}
%\end{table}
%\begin{table}[!hbp]
%\caption{Comparisons of different neuron sizes on NTU dataset (cross subject)}
%\label{table:resultNeuronSzieNTU}
%\centering
%\begin{tabular}{c|c|c|c|c|c|c|c}
%\hline
%$d$            &    64      &     80    &      100    &  128      &  150   &       200       &   300  \\
%\hline
%Accuracy (\%)  &  65.9      &    66.6    &    67.3    &  69.2    &   68.4  &       68.3      &  68.3    \\
%\hline
%\end{tabular}
%\end{table}
\begin{figure}
\begin{minipage}[b]{1.0\linewidth}
  \centering
  \centerline{\includegraphics[scale=.57]{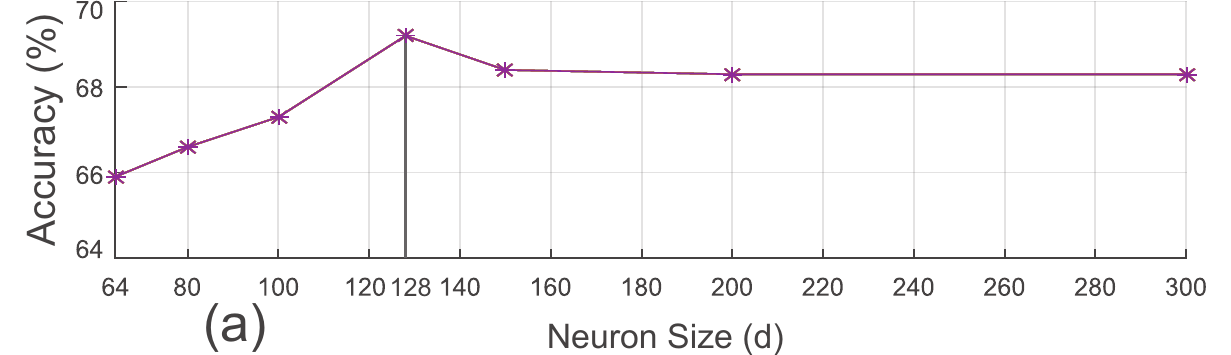}}
 % \centerline{(a)}
  \centerline{\includegraphics[scale=.57]{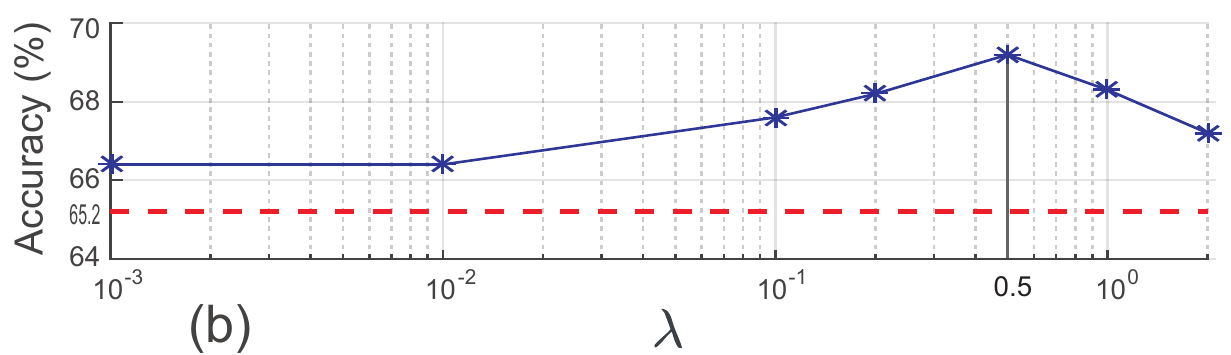}}
  %\centerline{(b)}
\end{minipage}
\caption{(a) Performance comparison of our approach using different values of neuron size ($d$) on the NTU RGB+D dataset (X-subject).
(b) Performance comparison of our method using different $\lambda$ values on the NTU RGB+D dataset (X-subject).
The blue line represents our results when different $\lambda$ values are used for trust gate,
while the red dashed line indicates the performance of our method when trust gate is not added.}
\label{fig:NTUResultLambda}
\end{figure}

Finally, we investigate the recognition performance with early stopping conditions
by feeding the first $p$ portion of the testing video to the trained network based on the cross-subject protocol ($p \in \{0.1, 0.2, ..., 1.0\}$).
The results are shown in \figurename{ \ref{fig:NTUResultEarlyStop}}.
We can observe that the results are improved when a larger portion of the video is fed to our network.

\begin{figure}
\begin{minipage}[b]{1.0\linewidth}
  \centering
  \centerline{\includegraphics[scale=.57]{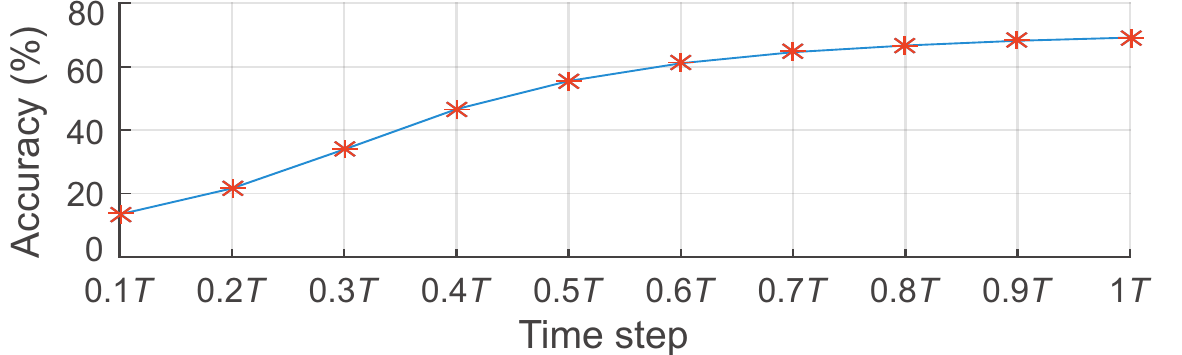}}
\end{minipage}
\caption{Experimental results of our method by early stopping the network evolution at different time steps.}
\label{fig:NTUResultEarlyStop}
\end{figure}

%Finally, we also compare the performance when different tree traversals are used. If we replace our tree traversal with top-down tree structure in our model, the accuracy drops to 46.5\% on NTU dataset (cross subject). It indicates that, as our tree traversal tries to share the parents and children information over the tree structure, it provides better performance than unidirectional tree traversal.

%We also evaluate the ``ST-LSTM (Bidirectional Joint Chain) + Trust Gate'' model on NTU dataset, and the accuracy is 67.2\% (cross subject), which is lower than 69.2\% achieved by ``ST-LSTM (Tree Traversal ) + Trust Gate'' model. It indicates our model with the new tree traversal strategy outperforms the simple joint chain based method and bidirectional joint chain model.

\subsection{Experiments on the UT-Kinect Dataset}
\label{sec:exp:resUTKinect}
%{\bf UT-Kinect Dataset.}
There are two evaluation protocols for the UT-Kinect dataset in the literature.
The first is the leave-one-out-cross-validation (LOOCV) protocol \cite{HOJ3D}.
The second protocol is suggested by \cite{zhu2013fusing}, for which half of the subjects are used for training, and the remaining are used for testing.
We evaluate our approach using both protocols on this dataset.

Using the LOOCV protocol,
our method achieves better performance than other skeleton-based methods,
as shown in \tablename{ \ref{table:resultUTKinectprotocol1}}.
Using the second protocol (see \tablename{ \ref{table:resultUTKinectprotocol2}}),
our method achieves competitive result (95.0\%) to the Elastic functional coding method \cite{anirudh2015elastic} (94.9\%),
which is an extension of the Lie Group model \cite{vemulapalli2014liegroup}.

\begin{table}[!htp]
		\caption{Experimental results on the UT-Kinect dataset (LOOCV protocol \cite{HOJ3D})}
		\label{table:resultUTKinectprotocol1}
		\centering
%\scriptsize
		\begin{tabular}{|l|c|c|}
			\hline
			Method & Feature & Acc.  \\
			\hline
			%\hline
			Grassmann Manifold \cite{slama2015accurate}  & Geometric & 88.5\%  \\
            Jetley \etal \cite{jetley20143d} & Geometric& 90.0\% \\
            Histogram of 3D Joints \cite{HOJ3D}  & Geometric & 90.9\%  \\
            Space Time Pose \cite{devanne2013space}  & Geometric & 91.5\%  \\
			Riemannian Manifold \cite{devanne20153d}  & Geometric & 91.5\% \\
            SCs (Informative Joints) \cite{jiang2015informative}  & Geometric & 91.9\% \\
            Chrungoo \etal \cite{chrungoo2014activity}  & Geometric & 92.0\%  \\
            Key-Pose-Motifs Mining\cite{Wang_2016_CVPR_Mining} & Geometric & 93.5\%  \\
            %Slama \etal \cite{slama2014grassmannian}  & Geometric & 95.3\% \\
			\hline
			ST-LSTM (Joint Chain)  & Geometric & 91.0\%  \\
			ST-LSTM (Tree) & Geometric  & 92.4\%   \\
			ST-LSTM (Tree) + Trust Gate  & Geometric & \textbf{97.0\%}  \\
			\hline
		\end{tabular}
\end{table}

\begin{table}[!htp]
		\caption{Results on the UT-Kinect dataset (half-vs-half protocol \cite{zhu2013fusing})}
		\label{table:resultUTKinectprotocol2}
		\centering
%\scriptsize
		\begin{tabular}{|l|c|c|}
			\hline
			Method & Feature & Acc.  \\
			\hline
            %\hline
			Skeleton Joint Features \cite{zhu2013fusing}   & Geometric & 87.9\%  \\
            Chrungoo \etal \cite{chrungoo2014activity}  & Geometric & 89.5\%  \\
			Lie Group \cite{vemulapalli2014liegroup} (reported by \cite{anirudh2015elastic})   & Geometric & 93.6\%   \\
			Elastic functional coding \cite{anirudh2015elastic}   & Geometric & 94.9\%   \\
			%ConvNets \cite{wang2015convnets} & 90.9\% \\
			\hline
			ST-LSTM (Tree) + Trust Gate   & Geometric & \textbf{95.0\%}  \\
			\hline
		\end{tabular}
\end{table}

Some actions in the UT-Kinect dataset involve human-object interactions, thus appearance based features representing visual information of the objects can be complementary to the geometric features.
Thus we can evaluate our proposed feature fusion approach within the ST-LSTM unit on this dataset.
The results are shown in \tablename{ \ref{table:resultUTFusion}.
Using geometric features only, the accuracy is 97\%.
By simply concatenating the geometric and visual features, the accuracy improves slightly.
However, the accuracy of our approach can reach 98\% when the proposed feature fusion method is adopted.

\begin{table}[h]
		\caption{Evaluation of our approach for feature fusion on the UT-Kinect dataset (LOOCV protocol \cite{HOJ3D}).
``Geometric + Visual'' indicates we simply concatenate the two types of features as the input.
``Geometric $\bigoplus$ Visual'' means we use the newly proposed feature fusion scheme within the ST-LSTM unit.}
		\label{table:resultUTFusion}
%\scriptsize
		\centering
		\begin{tabular}{|l|c|c|}
			\hline
			Feature Fusion Method & Acc. \\
			%\hline
			\hline
			Geometric Only         &  97.0\% \\
            Geometric + Visual    & 97.5\% \\
            Geometric $\bigoplus$ Visual  &98.0\%    \\
			\hline
		\end{tabular}
\\
\scriptsize
\end{table}

%\emph{NOTE: Lie Group paper \cite{vemulapalli2014liegroup}. The extension of lie group \cite{anirudh2015elastic}. For Table \ref{table:resultUTKinectprotocol2}: As lie group reported very high accuracy on UT-kinect dataset (97.2), so do we need to describe why in the table we choose the data from \cite{anirudh2015elastic}? Refer to section 4.2 of paper \cite{anirudh2015elastic}: FTP is a powerful tool to work around alignment issues ........}

\subsection{Experiments on the SBU Interaction Dataset}
\label{sec:exp:resSBU}
%{\bf SBU Interaction Dataset.}
We follow the standard evaluation protocol in \cite{yun2012two} and perform 5-fold cross validation on the SBU Interaction dataset.
As two human skeletons are provided in each frame of this dataset,
our traversal scheme visits the joints throughout the two skeletons over the spatial steps.

We report the results in terms of average classification accuracy in \tablename{ \ref{table:resultSBU}}.
The methods in \cite{zhu2016co} and \cite{du2015hierarchical} are both LSTM-based approaches, which are more relevant to our method.

\begin{table}[h]
		\caption{Experimental results on the SBU Interaction dataset}
		\label{table:resultSBU}
		\centering
%\scriptsize
		\begin{tabular}{|l|c|c|}
			\hline
			Method & Feature & Acc. \\
			%\hline
            \hline
			Yun \etal \cite{yun2012two} & Geometric & 80.3\% \\
			Ji \etal \cite{ji2014interactive} & Geometric & 86.9\%  \\
			CHARM \cite{li2015category} & Geometric & 83.9\%  \\
			Hierarchical RNN \cite{du2015hierarchical} & Geometric & 80.4\%  \\
			Co-occurrence LSTM \cite{zhu2016co} & Geometric & 90.4\%  \\
			Deep LSTM \cite{zhu2016co} & Geometric & 86.0\%  \\
            %Deep LSTM + Dropout \cite{zhu2016co} & Geometric & 89.7\%  \\
            %Deep LSTM + In-depth Dropout \cite{zhu2016co} & Geometric & 90.1\%  \\
            \hline
			ST-LSTM (Joint Chain) & Geometric & 84.7\%   \\
			ST-LSTM (Tree) & Geometric & 88.6\%   \\
			ST-LSTM (Tree) + Trust Gate & Geometric & \textbf{93.3\%}  \\
			\hline
		\end{tabular}		
\end{table}

The results show that the proposed ``ST-LSTM (Tree) + Trust Gate'' model outperforms all other skeleton-based methods.
``ST-LSTM (Tree)'' achieves higher accuracy than ``ST-LSTM (Joint Chain)'',
as the latter adds some false links between less related joints.
%We also evaluate the ``ST-LSTM (Double Joint Chain)'' on the SBU Interaction dataset,
%and its accuracy is 86.1\%.
%It is better than 84.7\% of ``ST-LSTM (Joint Chain)'',
%while worse than 88.6\% of ``ST-LSTM (Tree)''.

Both Co-occurrence LSTM \cite{zhu2016co} and Hierarchical RNN \cite{du2015hierarchical} adopt the Svaitzky-Golay filter \cite{savitzky1964smoothing} in the temporal domain
to smooth the skeletal joint positions and reduce the influence of noise in the data collected by Kinect.

The proposed ``ST-LSTM (Tree)'' model without the trust gate mechanism outperforms Hierarchical RNN,
and achieves comparable result (88.6\%) to Co-occurrence LSTM.
When the trust gate is used, the accuracy of our method jumps to 93.3\%.

\subsection{Experiments on the SYSU-3D Dataset}
\label{sec:exp:resSYSU}
%{\bf SYSU-3D Dataset.}

We follow the standard evaluation protocol in \cite{jianfang_CVPR15} on the SYSU-3D dataset.
The samples from 20 subjects are used to train the model parameters,
and the samples of the remaining 20 subjects are used for testing.
We perform 30-fold cross validation and report the mean accuracy in \tablename{~\ref{table:resultSYSU}}.

\begin{table}[h]
		\caption{Experimental results on the SYSU-3D dataset}
		\label{table:resultSYSU}
		\centering
%\scriptsize
		\begin{tabular}{|l|c|c|}
			\hline
			Method & Feature & Acc.  \\
			\hline
			%\hline
            LAFF (SKL) \cite{hu2016ECCV} & Geometric &  54.2\%  \\
            Dynamic Skeletons \cite{jianfang_CVPR15} & Geometric &  75.5\%  \\
			\hline
			%\hline
			ST-LSTM (Joint Chain)  & Geometric &  72.1\%  \\
			ST-LSTM (Tree) & Geometric  &  73.4\%   \\
			ST-LSTM (Tree) + Trust Gate  & Geometric & \textbf{76.5\%}  \\
			\hline
		\end{tabular}
\end{table}

The results in \tablename{~\ref{table:resultSYSU}} show that our proposed ``ST-LSTM (Tree) + Trust Gate'' method outperforms all the baseline methods on this dataset.
We can also find that the tree traversal strategy can help to improve the classification accuracy of our model.
As the skeletal joints provided by Kinect are noisy in this dataset,
the trust gate, which aims at handling noisy data, brings significant performance improvement (about 3\% improvement).

%We also test the performance of ``ST-LSTM (Double Joint Chain)'' on the dataset,
%and its accuracy is 72.5\%,
%which is slightly higher than 72.1\% achieved by ``ST-LSTM (Joint Chain)'',
%yet still lower than 73.4\% achieved by ``ST-LSTM (Tree)''.

%We investigate the effect of the last-to-first link scheme on the dataset.
%The results are shown in \tablename{~\ref{table:resultLinkSYSU}}.
%We observe that the recognition accuracy drops when the last-to-first link in our model is discarded.

%\begin{table}[h]
%\caption{Evaluation for last-to-first link on the SYSU-3D dataset}
%\label{table:resultLinkSYSU}
%\centering
%%\scriptsize
%\begin{tabular}{|l|c|}
%\hline
%Method & Acc.  \\
%\hline
%%\hline
%ST-LSTM (Tree) + Trust Gate (+LTF) &    76.5\% \\
%ST-LSTM (Tree) + Trust Gate (--LTF) & 	 75.9\%  \\
%\hline
%\end{tabular}
%\\
%\end{table}

There are large viewpoint variations in this dataset.
To make our model reliable against viewpoint variations,
we adopt a similar skeleton normalization procedure as suggested by \cite{nturgbd} on this dataset.
In this preprocessing step, we perform a rotation transformation on each skeleton,
such that all the normalized skeletons face to the same direction.
Specifically, after rotation, the 3D vector from ``right shoulder'' to ``left shoulder'' will be parallel to the X axis,
and the vector from ``hip center'' to ``spine'' will be aligned to the Y axis
(please see \cite{nturgbd} for more details about the normalization procedure).

We evaluate our ``ST-LSTM (Tree) + Trust Gate'' method by respectively using the original skeletons without rotation and the transformed skeletons,
and report the results in \tablename{~\ref{table:resultSYSURotation}}.
The results show that it is beneficial to use the transformed skeletons as the input for action recognition.

\begin{table}[h]
\caption{Evaluation for skeleton rotation on the SYSU-3D dataset}
\label{table:resultSYSURotation}
\centering
%\scriptsize
\begin{tabular}{|l|c|}
\hline
Method & Acc.  \\
\hline
%\hline
With Skeleton Rotation &        76.5\%    \\
Without Skeleton Rotation & 	73.0\%    \\
\hline
\end{tabular}
\\
\end{table}

\subsection{Experiments on the ChaLearn Gesture Dataset}
\label{sec:exp:resChaLearn}
%{\bf ChaLearn (2013) Gesture Dataset.}

%We follow the evaluation protocol adopted in \cite{fernando2015modeling,du2016representation} !!!!!!!!!!!!!!!!!!!!!!!!!!!!!!!!!!!
We follow the evaluation protocol adopted in \cite{wang2015hierarchical,fernando2015modeling}
and report the F1-score measures on the validation set of the ChaLearn Gesture dataset.

\begin{table}[h]
		\caption{Experimental results on the ChaLearn Gesture dataset}
		\label{table:resultChaLearn}
		\centering
%\scriptsize
		\begin{tabular}{|l|c|c|}
			\hline
			Method & Feature & F1-Score  \\
			\hline
			%\hline
             Portfolios \cite{yao2014gesture} & Geometric &  56.0\%  \\
             Wu \etal \cite{wu2013fusing} & Geometric &  59.6\%  \\
             Pfister \etal \cite{pfister2014domain} & Geometric &  61.7\%  \\
             HiVideoDarwin \cite{wang2015hierarchical} & Geometric &  74.6\%  \\
             VideoDarwin \cite{fernando2015modeling} & Geometric &  75.2\%  \\
             %\cite{} & Geometric &  \%  \\ \hline
             Deep LSTM \cite{nturgbd} & Geometric & 87.1\% \\
             %Du \etal \cite{du2016representation} & Geometric &  92.0\%  \\ !!!!!!!!!!!!!!!!!!!!!!!!!!!!!!!!!!!
			\hline
			%\hline
            ST-LSTM (Joint Chain)  & Geometric & 89.1\%  \\
            ST-LSTM (Tree) & Geometric & 89.9\%  \\
			ST-LSTM (Tree) + Trust Gate  & Geometric & \textbf{92.0\%}  \\
			\hline
		\end{tabular}
\end{table}

As shown in \tablename{~\ref{table:resultChaLearn}},
%our method achieves slightly superior performance over the deep learning based method in \cite{du2016representation},
our method surpasses the state-of-the-art methods \cite{yao2014gesture,wu2013fusing,pfister2014domain,wang2015hierarchical,fernando2015modeling,nturgbd},
which demonstrates the effectiveness of our method in dealing with skeleton-based action recognition problem.

Compared to other methods, our method focuses on modeling both temporal and spatial dependency patterns in skeleton sequences.
Moreover, the proposed trust gate is also incorporated to our method to handle the noisy skeleton data captured by Kinect,
which can further improve the results.

%The ``LSTM'' and ``LSTM + Trust Gate'' models are also evaluated on ChaLearn Gesture dataset.
%Their performance is 86.8\% and 87.6\%, respectively, which is lower than that of our proposed method (92.0\%).
%We also test the baseline of temporal average of the features on this dataset, and its performance is relatively low (\%).

\subsection{Experiments on the MSR Action3D Dataset}
\label{sec:exp:resMSR3D}
%{\bf MSR Action3D Dataset.}

We follow the experimental protocol in \cite{du2015hierarchical} on the MSR Action3D dataset,
and show the results in \tablename{~\ref{table:resultMSR3D}}.

On the MSR Action3D dataset, our proposed method, ``ST-LSTM (Tree) + Trust Gate'', achieves 94.8\% of classification accuracy,
which is superior to the Hierarchical RNN model \cite{du2015hierarchical} and other baseline methods.
%Besides the aforementioned datasets, we have also tested our model on the {\bf MSR Action3D dataset} \cite{li2010action} following the protocol in \cite{du2015hierarchical}, and achieved an accuracy of 94.8\%, which is slightly better than Hierarchical RNN \cite{du2015hierarchical} (94.5\%).

\begin{table}[h]
		\caption{Experimental results on the MSR Action3D dataset}
		\label{table:resultMSR3D}
		\centering
%\scriptsize
		\begin{tabular}{|l|c|c|}
			\hline
			Method & Feature & Acc.  \\
			\hline
			%\hline
            Histogram of 3D Joints \cite{HOJ3D}  & Geometric & 79.0\%  \\
            Joint Angles Similarities \cite{hog2-ohnbar}  & Geometric & 83.5\%  \\
            SCs (Informative Joints) \cite{jiang2015informative}  & Geometric & 88.3\% \\
			Oriented Displacements \cite{gowayyed2013histogram}  & Geometric & 91.3\%  \\
			Lie Group \cite{vemulapalli2014liegroup}  & Geometric & 92.5\% \\
            Space Time Pose \cite{devanne2013space}  & Geometric & 92.8\%  \\
            Lillo \etal \cite{lillo2016hierarchical}  & Geometric & 93.0\% \\
			Hierarchical RNN \cite{du2015hierarchical}  & Geometric & 94.5\%  \\
			\hline
			ST-LSTM (Tree) + Trust Gate  & Geometric & \textbf{94.8\%}  \\
			\hline
		\end{tabular}
\end{table}

\subsection{Experiments on the Berkeley MHAD Dataset}
\label{sec:exp:resMHAD}
%{\bf Berkeley MHAD Dataset.}
\begin{table}[h]
		\caption{Experimental results on the Berkeley MHAD dataset}
		\label{table:resultMHAD}
		\centering
%\scriptsize
		\begin{tabular}{|l|c|c|}
			\hline
			Method & Feature & Acc.  \\
			\hline
			%\hline
			Ofli \etal \cite{Ofli2014jvci}  & Geometric & 95.4\% \\
            Vantigodi \etal \cite{vantigodi2013real}  & Geometric & 96.1\%  \\
			Vantigodi \etal \cite{vantigodi2014action}  & Geometric & 97.6\%  \\
			Kapsouras \etal \cite{kapsouras2014action}  & Geometric & 98.2\%  \\
			Hierarchical RNN \cite{du2015hierarchical}  & Geometric & 100\%  \\
            Co-occurrence LSTM \cite{zhu2016co} & Geometric & 100\%  \\
			\hline
			ST-LSTM (Tree) + Trust Gate  & Geometric & \textbf{100\%}  \\
			\hline
		\end{tabular}
\end{table}

We adopt the experimental protocol in \cite{du2015hierarchical} on the Berkeley MHAD dataset.
384 video sequences corresponding to the first seven persons are used for training,
and the 275 sequences of the remaining five persons are held out for testing.
The experimental results in \tablename{ \ref{table:resultMHAD}} show that our method achieves very high accuracy (100\%) on this dataset.
Unlike \cite{du2015hierarchical} and \cite{zhu2016co}, our method does not use any preliminary manual smoothing procedures.

\subsection{Visualization of Trust Gates}
\label{sec:visualization}

%Adding noise to the input data does not work in our evaluation. However, using dropout can help to improve the performance. We set the dropout probability to 0.5 for SBU, CMU, Berkeley, UT-Kinect datasets, as they are relatively small. We found dropout just helps a little on NTU dataset.

In this section, to better investigate the effectiveness of the proposed trust gate scheme, we study the behavior of the proposed framework against the presence of noise in skeletal data from the MSR Action3D dataset.
We manually rectify some noisy joints of the samples by referring to the corresponding depth images.
We then compare the activations of trust gates on the noisy and rectified inputs.
As illustrated in \figurename{ \ref{fig:TrustGateEffect}(a)},
the magnitude of trust gate's output ($l_2$ norm of the activations of the trust gate) is smaller when a noisy joint is fed, compared to the corresponding rectified joint.
This demonstrates how the network controls the impact of noisy input on its stored representation of the observed data.

In our next experiment, we manually add noise to one joint for all testing samples on the Berkeley MHAD dataset, in order to further analyze the behavior of our proposed trust gate.
Note that the Berkeley MHAD dataset was collected with motion capture system, thus
the skeletal joint coordinates in this dataset are much more accurate than those captured with Kinect sensors.
%So this dataset is quite fit to evaluate the performance of the trust gate if some manually noise is added.

We add noise to the right foot joint by moving the joint away from its original location.
The direction of the translation vector is randomly chosen and the norm is a random value around $30cm$, which is a significant noise in the scale of human body.
%For each video, we add noise to the same joint at the same time step, and then analyse the effect in average.
We measure the difference in the magnitudes of trust gates' activations between the noisy data and the original ones.
For all testing samples, we carry out the same operations and then calculate the average difference.
The results in \figurename{ \ref{fig:TrustGateEffect}(b)} show that the magnitude of trust gate is reduced when the noisy data is fed to the network.
This shows that our network tries to block the flow of noisy input and stop it from affecting the memory.
We also observe that the overall accuracy of our network does not drop after adding the above-mentioned noise to the input data.

\begin{figure}[htb]
\begin{minipage}[b]{0.47\linewidth}
  \centering
  \centerline{\includegraphics[scale=.53]{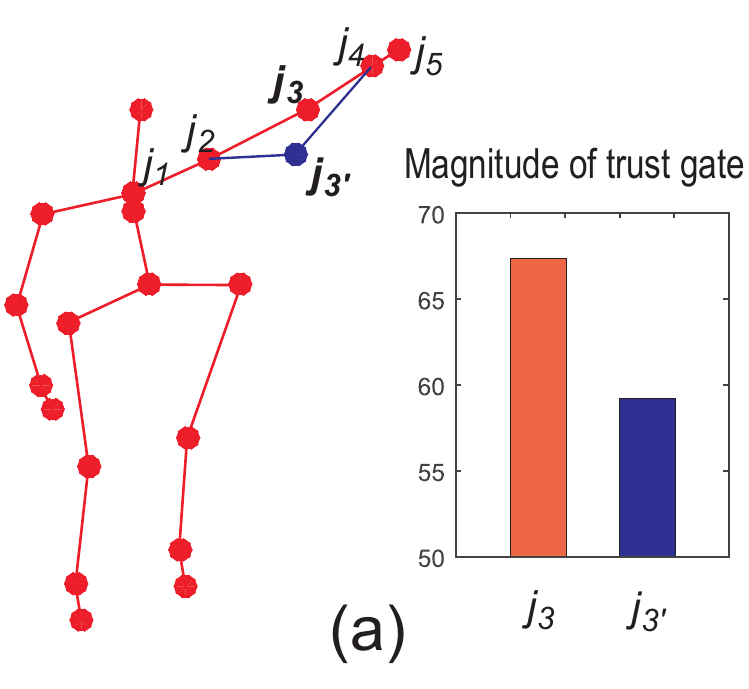}}
\end{minipage}
\begin{minipage}[b]{0.52\linewidth}
  \centering
  \centerline{\includegraphics[scale=.53]{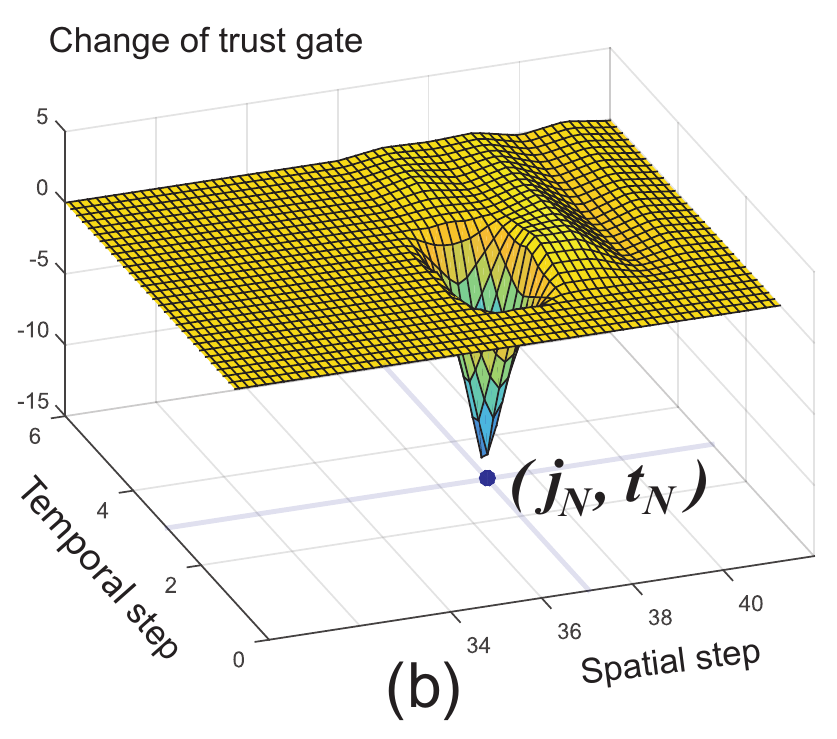}}
\end{minipage}
%\begin{minipage}[b]{1.0\linewidth}
%  \centering
%  \centerline{\includegraphics[scale=.25,trim=0 180 0 180,clip]{trustgate_mhad2.pdf}}
%\end{minipage}
\caption{Visualization of the trust gate's behavior when inputting noisy data.
(a) $j_{3'}$ is a noisy joint position, and $j_3$ is the corresponding rectified joint location.
In the histogram, the blue bar indicates the magnitude of trust gate when inputting the noisy joint $j_{3'}$.
The red bar indicates the magnitude of the corresponding trust gate when $j_{3'}$ is rectified to $j_3$.
(b) Visualization of the difference between the trust gate calculated when the noise is imposed at the step $(j_N, t_N)$ and that calculated when inputting the original data.}
\label{fig:TrustGateEffect}
\end{figure}

\begin{table*}[htb]
\caption{Performance comparison of different spatial sequence models}
\label{table:resultDoubleChain}
\centering
\footnotesize
\begin{tabular}{|c|c|c|c|c|c|}
\hline
~~~~~~~~~~~~~~~~~~Dataset~~~~~~~~~~~~~~~~~~ & NTU (X-Subject) & NTU (X-View)  & ~~~UT-Kinect~~~ & SBU Interaction & ChaLearn Gesture \\
\hline
%\hline
ST-LSTM (Joint Chain)          &   61.7\%  &   75.5\%  &   91.0\%     &   84.7\%   &    89.1\%  \\
ST-LSTM (Double Joint Chain)   &   63.5\%  &    75.6\%  &   91.5\%    &   85.9\%   &    89.2\%  \\
ST-LSTM (Tree)                 &   65.2\%  &    76.1\% &   92.4\%    &    88.6\%  &    89.9\%  \\
\hline
\end{tabular}
\\
\end{table*}

\begin{table*}[tb]
\caption{Performance comparison of Temporal Average, LSTM, and our proposed ST-LSTM}
\label{table:resultLSTMTG}
\centering
\footnotesize
\begin{tabular}{|c|c|c|c|c|c|}
\hline
~~~~~~~~~~~~~~~~~~Dataset~~~~~~~~~~~~~~~~~~   & NTU (X-Subject)         & NTU (X-View)     & ~~~UT-Kinect~~~   & SBU Interaction            & ChaLearn Gesture\\
\hline
%\hline
Temporal Average                       &       47.6\%       &              52.6\%    &   81.9\%             &   71.5\%              &        77.9\%   \\
\hline
LSTM                      &   62.0\%            &   70.7\%           &   90.5\%              &   86.0\%                  &        87.1\%      \\
LSTM + Trust Gate         &   62.9\%           &   71.7\%             &   92.0\%             &   86.6\%                  &        87.6\%      \\
\hline
ST-LSTM                  &   65.2\%          &    76.1\%               &   92.4\%              &    88.6\%               &         89.9\%     \\
ST-LSTM + Trust Gate    &    69.2\%          &     77.7\%              &    97.0\%         &        93.3\%               &    92.0\%           \\
\hline
\end{tabular}
\\
\end{table*}

\begin{table*}[tb]
\caption{Evaluation of the last-to-first link in our proposed network}
\label{table:resultLTFLink}
\centering
\footnotesize
\begin{tabular}{|c|c|c|c|c|c|}
\hline
~~~~~~~~~~~~~~~~~~Dataset~~~~~~~~~~~~~~~~~~ & NTU (X-Subject) & NTU (X-View) & ~~~UT-Kinect~~~ & SBU Interaction & ChaLearn Gesture \\
\hline
%\hline
Without last-to-first link    &   68.5\% &     76.9\%     &   96.5\%     &   92.1\%   &  90.9  \%  \\
With last-to-first link       &   69.2\% &     77.7\%     &   97.0\%    &    93.3\%  &   92.0 \%  \\
\hline
\end{tabular}
\\
\end{table*}

%TODO: AN OPENING SENTENCE IS MISSING HERE. WE HAVE TO OPEN THE SECTION WITH A KEY SENTENCE INTRODUCING WHAT WE ARE GOING TO DO HERE... PLEASE TRY TO ADD A SENTENCE HERE...
%IT SEEMS THE TITLE OF THE SUBSECTION IS NOT SUITABLE EITHER... PLEASE TRY TO FIND A REASONABLE NAME SIMILAR TO 4.10
%
%IN DISCUSSION SECTION WE NEED TO WRAP UP ALL OF OUR EXPERIMENTS AND DISCUSS WHAT WE HAVE ACHIEVED, BUT IT SEEMS SOME NEW EXPERIMENTS ARE INTRODUCED HERE... please revise this section first, i will revise it tomorrow...

\subsection{Evaluation of Different Spatial Joint Sequence Models}
\label{sec:discussion1}

The previous experiments showed how ``ST-LSTM (Tree)'' outperforms ``ST-LSTM (Joint Chain)'', because ``ST-LSTM (Tree)'' models the kinematic dependency structures of human skeletal sequences.
In this section, we further analyze the effectiveness of our ``ST-LSTM (Tree)'' model and compare it with a ``ST-LSTM (Double Joint Chain)'' model.

The ``ST-LSTM (Joint Chain)'' has fewer steps in the spatial dimension than the ``ST-LSTM (Tree)''.
One question that may rise here is if the advantage of ``ST-LSTM (Tree)'' model could be only due to the higher length and redundant sequence of the joints fed to the network, and not because of the proposed semantic relations between the joints.
To answer this question, we evaluate the effect of using a double chain scheme to increase the spatial steps of the ``ST-LSTM (Joint Chain)'' model.
Specifically, we use the joint visiting order of 1-2-3-...-16-1-2-3-...-16,
and we call this model as ``ST-LSTM (Double Joint Chain)''.
The results in \tablename{~\ref{table:resultDoubleChain}} show that the performance of ``ST-LSTM (Double Joint Chain)'' is better than ``ST-LSTM (Joint Chain)'',
yet inferior to ``ST-LSTM (Tree)''.

This experiment indicates that it is beneficial to introduce more passes in the spatial dimension to the ST-LSTM for performance improvement.
A possible explanation is that the units visited in the second round can obtain the global level context representation from the previous pass,
thus they can generate better representations of the action patterns by using the context information.
%These better representations can help to improve the classification accuracy of the sequence.
However, the performance of ``ST-LSTM (Double Joint Chain)'' is still weaker than ``ST-LSTM (Tree)'',
though the numbers of their spatial steps are almost equal.

The proposed tree traversal scheme is superior because it connects the most semantically related joints
and avoids false connections between the less-related joints (unlike the other two compared models).

\subsection{Evaluation of Temporal Average, LSTM and ST-LSTM}
\label{sec:discussion2}

To further investigate the effect of simultaneous modeling of dependencies in spatial and temporal domains,
in this experiment, we replace our ST-LSTM with the original LSTM which only models the temporal dynamics among the frames without explicitly considering spatial dependencies.
%We replace the ST-LSTM with the original LSTM,
%which models the temporal dynamics over the frames without explicitly considering spatial dependencies,
%for our network architecture to further investigate the effect of simultaneously modeling the dependencies in both domains.
We report the results of this experiment in \tablename{ \ref{table:resultLSTMTG}}.
As can be seen, our ``ST-LSTM + Trust Gate'' significantly outperforms ``LSTM + Trust Gate''.
This demonstrates that the proposed modeling of the dependencies in both temporal and spatial dimensions provides much richer representations than the original LSTM.

The second observation of this experiment is that if we add our trust gate to the original LSTM,
the performance of LSTM can also be improved,
but its performance gain is less than the performance gain on ST-LSTM.
A possible explanation is that we have both spatial and temporal context information at each step of ST-LSTM to generate a good prediction of the input at the current step ((see Eq. (\ref{eq:p_j_t})),
thus our trust gate can achieve a good estimation of the reliability of the input at each step by using the prediction (see Eq. (\ref{eq:tau})).
However, in the original LSTM, the available context at each step is from the previous temporal step,
i.e., the prediction can only be based on the context in the temporal dimension,
thus the effectiveness of the trust gate is limited when it is added to the original LSTM.
This further demonstrates the effectiveness of our ST-LSTM framework for spatio-temporal modeling of the skeleton sequences.

%The second observation of this experiment is the effect of inserting the proposed trust gate to the ordinary LSTM, for which it gains better performance.
%However, this improvement is more significant when the trust gate is utilized in the proposed ST-LSTM network (in contrast with ordinary LSTM).
%This advantage is due to the modeling of both spatial and temporal context information at each step of ST-LSTM which generates richer prediction of the input (see Eq. (\ref{eq:p_j_t}), and therefore a better estimation for the input reliability (see Eq. (\ref{eq:tau}), at each step.
%However, in the ordinary LSTM model, the context information at each step is from the previous temporal steps, which limits the trust gate's prediction to temporal domain context.
%This further demonstrates the effectiveness of our ST-LSTM framework for spatio-temporal modeling of the skeleton sequences.

In addition, we investigate the effectiveness of the LSTM structure for handling the sequential data.
We evaluate a baseline method (called ``Temporal Average'') by averaging the features from all frames instead of using LSTM.
Specifically, the geometric features are averaged over all the frames of the input sequence (i.e., the temporal ordering information in the sequence is ignored),
and then the resultant averaged feature is fed to a two-layer network, followed by a softmax classifier.
The performance of this scheme is much weaker than our proposed ST-LSTM with trust gate,
and also weaker than the original LSTM, as shown in \tablename{~\ref{table:resultLSTMTG}}.
The results demonstrate the representation strengths of the LSTM networks for modeling the dependencies and dynamics in sequential data, when compared to traditional temporal aggregation methods of input sequences.

\subsection{Evaluation of the Last-to-first Link Scheme}
\label{sec:discussion3}

In this section, we evaluate the effectiveness of the last-to-first link in our model (see section \ref{sec:approach:learning}).
The results in \tablename{ \ref{table:resultLTFLink}} show the advantages of using the last-to-first link in improving the final action recognition performance.

\section{Conclusion}
\label{sec:conclusion}
In this paper, we have extended the RNN-based action recognition method to both spatial and temporal domains.
Specifically, we have proposed a novel ST-LSTM network which analyzes the 3D locations of skeletal joints at each frame and at each processing step.
%In this paper, the RNN-based action recognition method over both spatial and temporal domains is studied.
%Specifically, we proposed a novel ST-LSTM network which analyzes the 3D locations of skeletal body joints at each frame and at each processing step.
%For better representation of the structured input to the network,
A skeleton tree traversal method based on the adjacency graph of body joints is also proposed to better represent the structure of the input sequences and
to improve the performance of our network by connecting the most related joints together in the input sequence.
%Due to the unreliability of the input data,
In addition, a new gating mechanism is introduced to improve the robustness of our network against the noise in input sequences.
A multi-modal feature fusion method is also proposed for our ST-LSTM framework.
The experimental results have validated the contributions and demonstrated the effectiveness of our approach
which achieves better performance over the existing state-of-the-art methods on seven challenging benchmark datasets.

\section*{Acknowledgement}
%The research is supported by Singapore Ministry of Education (MOE) Tier 2 ARC28/14, and Singapore A*STAR Science and Engineering Research Council PSF1321202099.
This work was carried out at Rapid-Rich Object Search (ROSE) Lab, Nanyang Technological University.
ROSE Lab is supported by the National Research Foundation, Singapore, under its IDM Strategic Research Programme.
\ifCLASSOPTIONcaptionsoff
  \newpage
\fi

% trigger a \newpage just before the given reference
% number - used to balance the columns on the last page
% adjust value as needed - may need to be readjusted if
% the document is modified later
%\IEEEtriggeratref{8}
% The "triggered" command can be changed if desired:
%\IEEEtriggercmd{\enlargethispage{-5in}}

% references section

% can use a bibliography generated by BibTeX as a .bbl file
% BibTeX documentation can be easily obtained at:
% http://mirror.ctan.org/biblio/bibtex/contrib/doc/
% The IEEEtran BibTeX style support page is at:
% http://www.michaelshell.org/tex/ieeetran/bibtex/
%\bibliographystyle{IEEEtran}
% argument is your BibTeX string definitions and bibliography database(s)
%\bibliography{IEEEabrv,../bib/paper}
%
% <OR> manually copy in the resultant .bbl file
% set second argument of \begin to the number of references
% (used to reserve space for the reference number labels box)

\bibliographystyle{IEEEtran}
\end{document}